# Generative artificial intelligence for reliable mechanistic reasoning for corrosion

Bharath M N[a,b,c], R K Singh Raman[b], Alankar Alankar[ade]
[a]Department of Mechanical Engineering, Indian Institute of Technology Bombay, Mumbai, Maharashtra, 400076, India
[b]Department of Mechanical and Aerospace Engineering, Monash University, Clayton 3800, Victoria, Australia
Email: raman.singh@monash.edu
[c]IITB-Monash Research Academy, Mumbai, Maharashtra, 400076, India, Email: alankar.alankar@iitb.ac.in
[d]Center for Machine Intelligence and Data Science, Indian Institute of Technology Bombay, Mumbai, Maharashtra, 400076, India

**Abstract**
Corrosion accounts for approximately 4% of global GDP, and reliable prediction is essential for timely mitigation. Machine learning effectively predicts corrosion rates from composition, microstructure, and environmental variables, but cannot explain the underlying mechanisms. A reliable approach in safety-critical materials engineering requires not only accurate retrieval but also mechanistically defensible reasoning, a capability that existing factuality metrics cannot assess. This work presents a domain-adapted retrieval-augmented generation framework for corrosion knowledge synthesis, demonstrated on magnesium alloy corrosion. Three open-weight language models (Llama-3.1-8B, Qwen-2.5-7B, Mistral-7B) are fine-tuned on 3,309 expert-verified question-answer pairs from 840 peer-reviewed papers and integrated with a hybrid dense-lexical retrieval pipeline. Retrieval augmentation produces Token F1 gains of 143–194%, with system faithfulness of 0.964 and context recall of 0.988. Blind external validation on newly published literature and in-house electrochemical data confirms trend-level generalisation. Reason Map, a proposition-graph framework, is further introduced; it independently constructs directed evidence graphs from generated answers and retrieved literature, enabling systematic detection of causal direction inversions and unsupported inferential leaps that flat factuality metrics cannot expose. The modular architecture can be applied across domains, offering a generalizable blueprint for trustworthy AI-assisted knowledge synthesis to circumvent corrosion, which can also be applied to other engineering domains.

## 1. Corrosion assessment in the age of data-driven methods: from pattern recognition to mechanistic reasoning

Artificial intelligence (AI) is beginning to show promise in understanding complex corrosion phenomena, including the adsorption of aggressive species and inhibitors on corroding surfaces.[1,2] This progress is important because corrosion remains a major global challenge, costing approximately US$2.5 trillion annually and affecting infrastructure, energy systems, transport networks, and biomedical devices worldwide.[3] Reliable assessment requires integrating electrochemical, chemical, and transport processes that interact simultaneously across alloy composition, microstructure, electrolyte chemistry, and exposure conditions. This combinatorial complexity is something conventional experimental methods, for all their

mechanistic rigour, cannot scale across.[3] This has motivated a sustained transition toward machine learning (ML) for corrosion rate prediction, structure-property modelling, and risk-informed design, with methods including gradient boosting, neural networks, and vision-based quantification now applied across multiple alloy systems.[4–7] Yet most deployed models operate within a knowledge-based paradigm wherein predictions are derived from patterns embedded in narrow, domain-specific datasets, without access to the mechanistic reasoning accumulated across decades of published corrosion science, limiting both generalisation and scientific interpretability (Table S1).[4,7]

Large language models (LLMs) coupled with retrieval-augmented generation (RAG) offer a qualitatively different capability: evidence-grounded synthesis across dispersed textual knowledge, instead of pattern interpolation within a fixed training distribution.[8,9] For corrosion science, where authoritative evidence is distributed across thousands of peer-reviewed papers, technical standards, and experimental databases, this raises the prospect of systems that assemble evidence across composition, environment, and degradation pathway to produce mechanistically meaningful explanations. Recent work has begun to establish the feasibility of this approach: domain-adapted transformers have improved literature-based candidate discovery and structured knowledge extraction,[10–14] fine-tuned LLMs with RAG have demonstrated competitive accuracy on corrosion-specific question answering,[9] and LLM-driven electrochemical impedance spectroscopy interpretation has achieved classification accuracies exceeding 90% on experimental coating datasets.[15] However, all existing systems share a critical limitation: none has been validated on literature published after training and indexing, the only test that distinguishes genuine generalisation from sophisticated memorisation. More fundamentally, retrieval does not guarantee sound reasoning. A model may retrieve relevant evidence yet still invert causal relationships, omit the inferential steps linking evidence to conclusions, or combine claims from mutually inconsistent sources. In safety-critical engineering, such failures can directly affect design, monitoring, and qualification decisions.

Existing evaluation frameworks are not equipped to detect these errors. Factuality metrics such as FActScore, SAFE, and RAGAS break a generated answer into atomic claims and verify each claim independently against retrieved evidence.[16–18] Knowledge-based approaches similarly test consistency against structured fact repositories, rather than the coherence of the underlying inference.[19,20] Both approaches share the same blind spot: they evaluate propositions one at a time and cannot see errors in the reasoning chain that links them. This matters even more in mechanistic domains. Figure 1 presents a representative case from this study, involving a generated answer on the effect of a rare-earth element on magnesium alloy corrosion. The answer achieved a RAGAS faithfulness score of 0.964, and seven of nine proposition nodes were individually supported by retrieved evidence, with entailment scores ranging from 0.919 to 0.998. Nonetheless, the response contained a causal inversion. It claimed that denser rare-earth-oxide surface films caused coherent growth along the MgO (002) plane, whereas the retrieved evidence supported the reverse causal relationship. The relevant propositions passed independent natural language inference (NLI) verification, making the error invisible to flat factuality metrics. It became detectable only by comparing the directed structure of the answer with an independently constructed evidence graph. In corrosion science, such structural errors are consequential. The direction and sequence of electrochemical and surface processes can determine whether a practitioner treats film nucleation mode or film density as the primary design variable.

Here we present an end-to-end framework for domain-adapted corrosion knowledge synthesis, validated across automatic metrics, blinded expert human assessment, blind external evaluation on newly published literature, and independent experimental electrochemical data. Three open-weight language models, Llama-3.1-8B instruct, Qwen-2.5-7B instruct, and Mistral-7B instruct, are fine-tuned via low-rank adaptation on 3,309 expert-verified question-answer (Q-A) pairs derived from 840 peer-reviewed papers in magnesium alloy corrosion, a mechanistically rich and safety-critical proof-of-concept domain, and integrated with a hybrid dense–lexical retrieval pipeline combining reciprocal rank fusion and cross-encoder reranking. We introduce Reason Map, a proposition-graph framework that compares the reasoning structure of generated answers against independently constructed evidence graphs from retrieved literature. This structural comparison enables systematic detection of causal inversions and unsupported inferential leaps that flat factuality metrics miss. Built on a modular, open-weight and domain-agnostic architecture, Reason Map offers a replicable blueprint for trustworthy AI-assisted knowledge synthesis in corrosion science and related engineering fields, where scientifically defensible reasoning has crucial practical consequences.

**(a)**

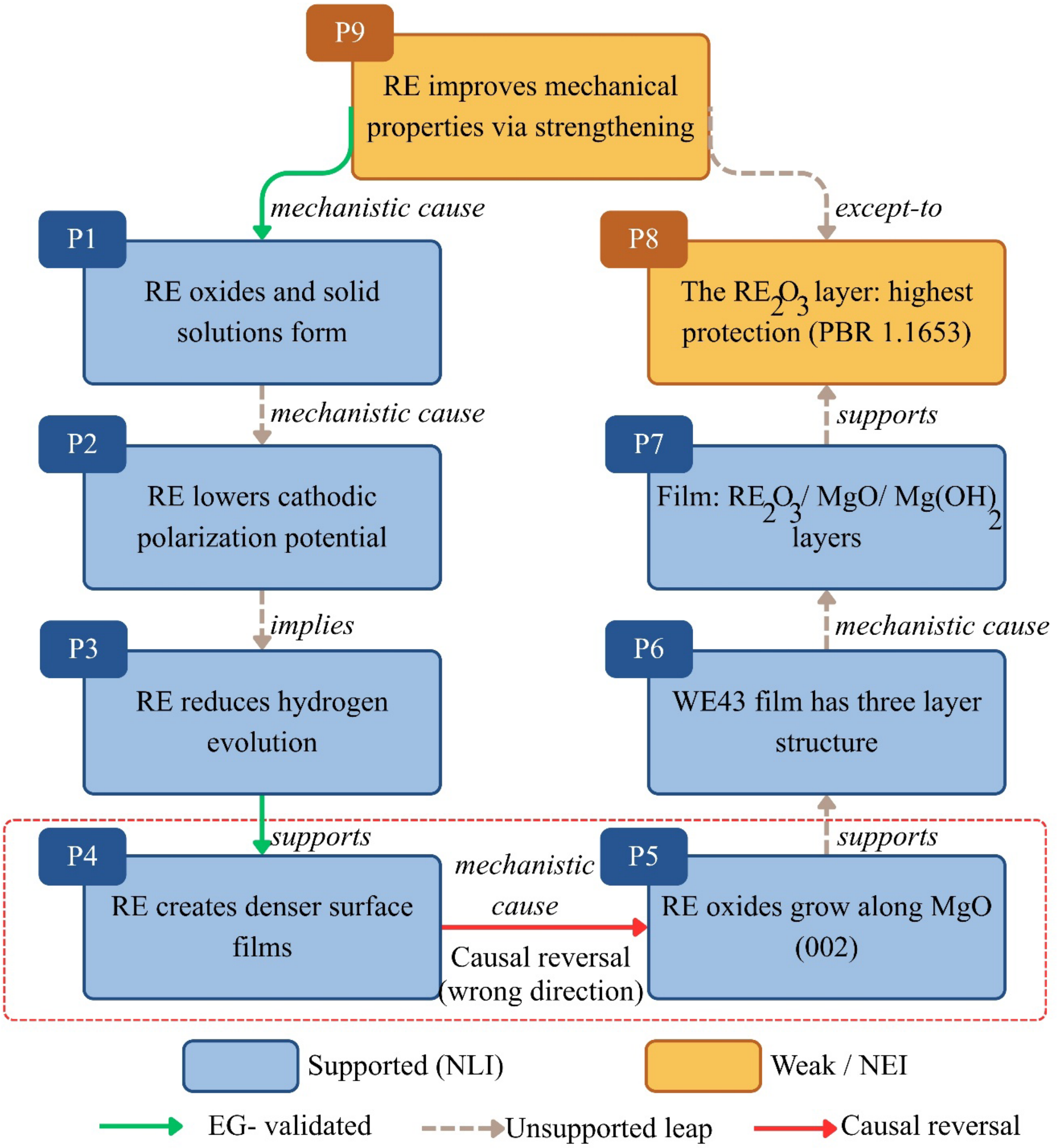
P9
RE improves mechanical properties via strengthening
mechanistic cause
except-to
P1
RE oxides and solid solutions form
P8
The $RE_2O_3$ layer: highest protection (PBR 1.1653)
mechanistic cause
supports
P2
RE lowers cathodic polarization potential
P7
Film: $RE_2O_3$/ MgO/ $Mg(OH)_2$ layers
implies
mechanistic cause
P3
RE reduces hydrogen evolution
P6
WE43 film has three layer structure
supports
supports
P4
RE creates denser surface films
mechanistic cause
Causal reversal (wrong direction)
P5
RE oxides grow along MgO (002)
Supported (NLI)
Weak / NEI
EG- validated
Unsupported leap
Causal reversal

(b)

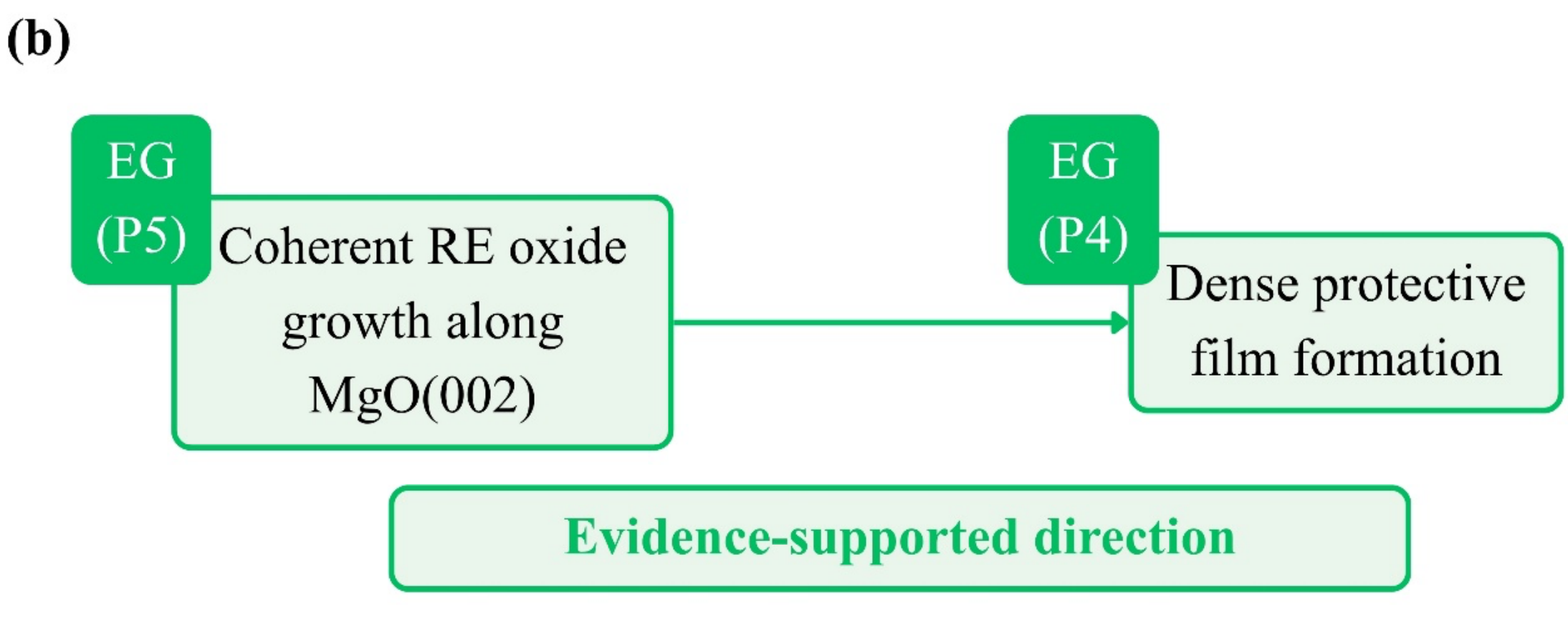

(c)

| | *Claim* | *Outgoing relation* | *NLI verdict* |
|---|---|---|---|
| P1 | Rare earth elements form stable RE oxides and solid solutions in Mg alloys. | *mechanistic cause* | ● supported |
| P2 | RE oxides lower cathodic polarization potential of Mg alloys. | *implies* | ● supported |
| P3 | RE additions reduce hydrogen evolution during Mg corrosion. | *supports* | ● supported |
| P4 | RE oxides create denser surface films on Mg alloys. | *mechanistic cause* | ● supported |
| P5 | RE oxides grow coherently along the (002) plane of MgO. | *supports* | ● supported |
| P6 | The corrosion product film on aged WE43 alloy is a three-layer structure. | *mechanistic cause* | ● supported |
| P7 | The three-layer film consists of $RE_2O_3$, MgO doped with $RE_2O_3$, and $Mg(OH)_2$ doped with $RE_2O_3$. | *mechanistic cause* | ● supported |
| P8 | The $RE_2O_3$ layer provides the highest protection with PBR of 1.1653. | - | ● weak / NEI |
| P9 | RE improves mechanical properties through second-phase hardening, solid solution strengthening, and precipitation strengthening. | *mechanistic cause* | ● weak / NEI |

**Figure 1. Reason Map detects a causal inversion missed by claim-level factuality checking.**

Reason Map output for the query "What is the role of rare-earth elements in Mg alloy corrosion?", generated by Llama-3.1-8B-Instruct with hybrid retrieval and Phase 1 + Phase 2 validation. **(a) Answer graph.** Nine proposition nodes, P1–P9, were extracted from the generated answer and connected by nine typed inference edges. Node colour indicates the natural language inference (NLI) verdict against retrieved evidence: blue indicates supported claims and amber indicates not enough information (NEI). Seven nodes were NLI-supported, with entailment scores ranging from 0.919 to 0.998, whereas P8 and P9 were NEI. Edge style indicates alignment with the independently constructed evidence graph (EG): solid green denotes an EG-validated relation, dashed grey denotes an unsupported inferential leap, and solid red denotes a causal reversal. The red dashed box marks the only causal reversal detected, P4→P5. The answer graph asserts that denser rare-earth-oxide surface films cause coherent growth along the MgO (002) plane, whereas the retrieved evidence supports the reverse direction. Two edges are EG-validated (P9→P1 and P3→P4), six are unsupported inferential leaps (P1→P2, P2→P3, P5→P6, P6→P7, P7→P8 and P9→P8), and one is a causal reversal (P4→P5). **(b) Evidence graph.** The evidence graph, built independently from the top five retrieved chunks without reference to the generated answer, contains the evidence-supported edge EG(P5)→EG(P4): coherent rare-earth-oxide growth along the MgO (002) plane as the mechanistic cause of dense protective film formation. This confirms that the answer-graph direction P4→P5 is inverted relative to the evidence. P4 and P5 both pass NLI verification individually, with entailment scores of 0.995 and 0.919, respectively; the causal error is therefore undetectable by flat claim-level checking and is exposed only by dual-graph alignment. **(c) Proposition key.** Claim text, outgoing inference relation and NLI verdict for each proposition node. Overall, seven of nine propositions were evidence-supported, two were NEI, no node was directly contradicted, one causal reversal was detected, and six of nine inference edges were unsupported by the evidence graph.

## 2. Results

### 2.1 Domain-adapted language models as generation baselines

Three open-weight language models, Llama-3.1-8B-Instruct, Qwen-2.5-7B-Instruct, and Mistral-7B-Instruct-v0.3, were fine-tuned using low-rank adaptation (LoRA, rank 16) on 2,604 Q-A pairs from the Magnesium Corrosion Knowledge Dataset (Table S2, S3), with performance evaluated on a held-out test set of 140 questions. All three models were fine-tuned under identical hyperparameter configurations to ensure that observed performance differences reflect architectural and pretraining characteristics rather than tuning variability.

Without retrieval augmentation, Llama-3.1-8B achieved the strongest baseline performance across all metrics, with a Token F1 of 0.635, ROUGE-L of 0.581, BLEU of 0.413, and BERTScore F1 of 0.924. Mistral-7B and Qwen-2.5-7B produced substantially lower lexical scores with Token F1values of 0.327 and 0.193, respectively, but retained moderate semantic alignment, as shown by BERTScore F1: 0.852 and 0.848. This pattern suggests that both models generated domain-appropriate responses at the semantic level, but struggled to reproduce the precise technical phrasing, alloy designations, corrosion mechanisms and experimental conditions that characterise expert reference answers. The BLEU gap was most pronounced, with Llama scoring 0.413 compared with 0.045 for Mistral and 0.022 for Qwen. This reflects BLEU's sensitivity to exact phrase matching, as well as the shorter average response lengths of Mistral and Qwen, at 64.3 and 32.1 tokens, respectively, compared with 87.2 tokens for Llama (Figure 2).

The LoRA fine-tuning dynamics of Llama-3.1-8B were monitored over five epochs, with early stopping based on validation Token F1. The model showed a stable learning trajectory, performance improved through the first three epochs as it captured domain-specific patterns, before stabilising across epochs 4-5. This plateau suggests that the model had largely saturated the available fine-tuning signal, rather than continuing into uncontrolled overfitting. Token F1 peaked at epoch 4 (0.630) and early stopping was triggered after epoch 5, with a final Token F1 of 0.628 on the validation set. The moderate final generalisation gap of 0.93 is consistent with domain-specific fine-tuning on a specialist corpus, reflecting a balance between training optimisation and generalisation capability (Figure 3, Table S4). These fine-tuned models, with their domain-adapted parametric knowledge, serve as the generation components of the RAG pipeline evaluated in the following sections.

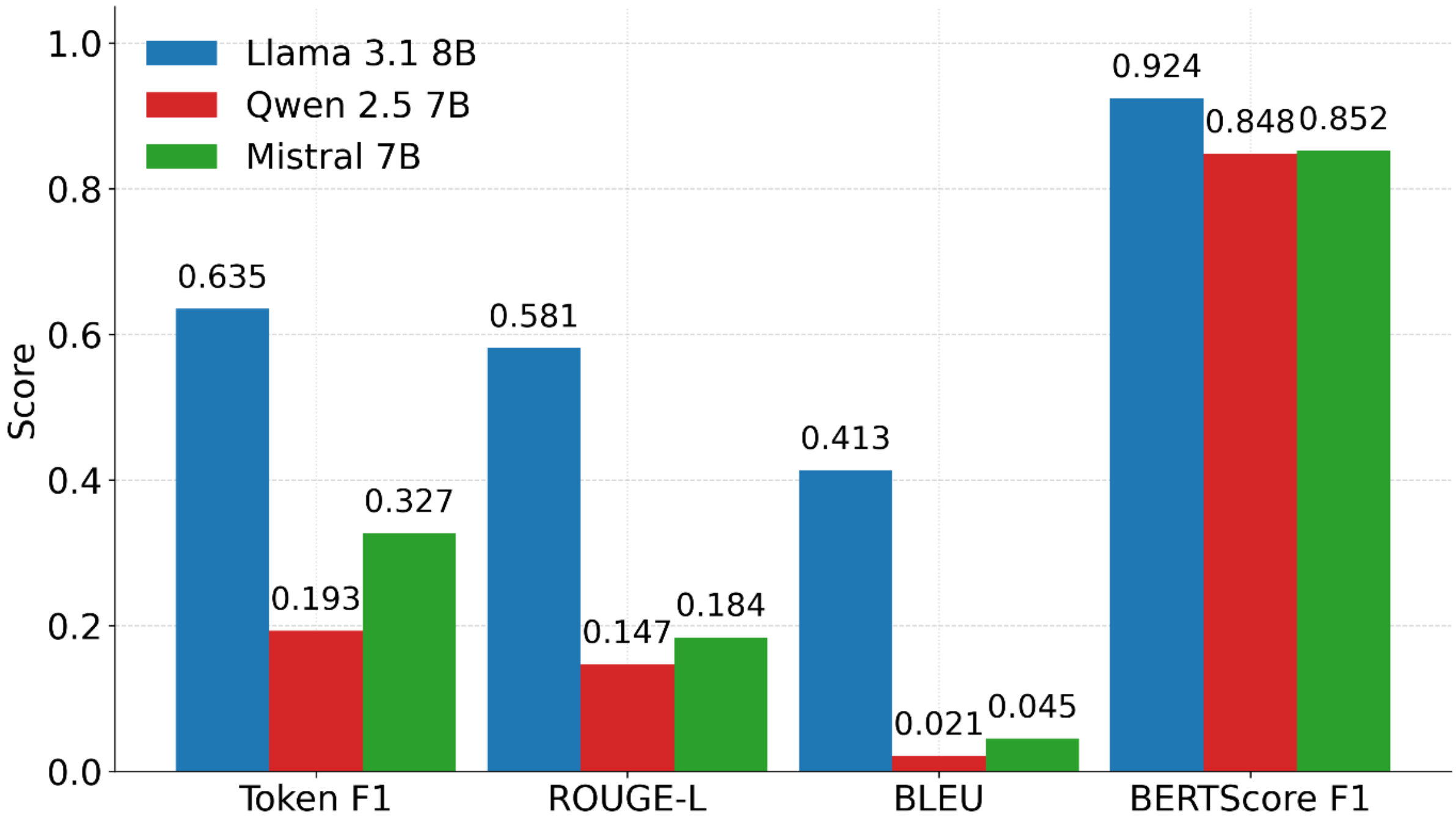

**Figure 2. Performance metrics comparison**
Llama 3.1 8B significantly outperforms Mistral 7B v0.3 and Qwen 2.5 7B across all evaluation metrics on test set ($N$=140). Llama's advantage is most pronounced for BLEU (9× vs Mistral, 43× vs Qwen), indicating superior $n$-gram matching for technical terminology. BERTScore gap is smallest (8% vs Mistral), revealing a semantic understanding-to-generation gap in smaller models.

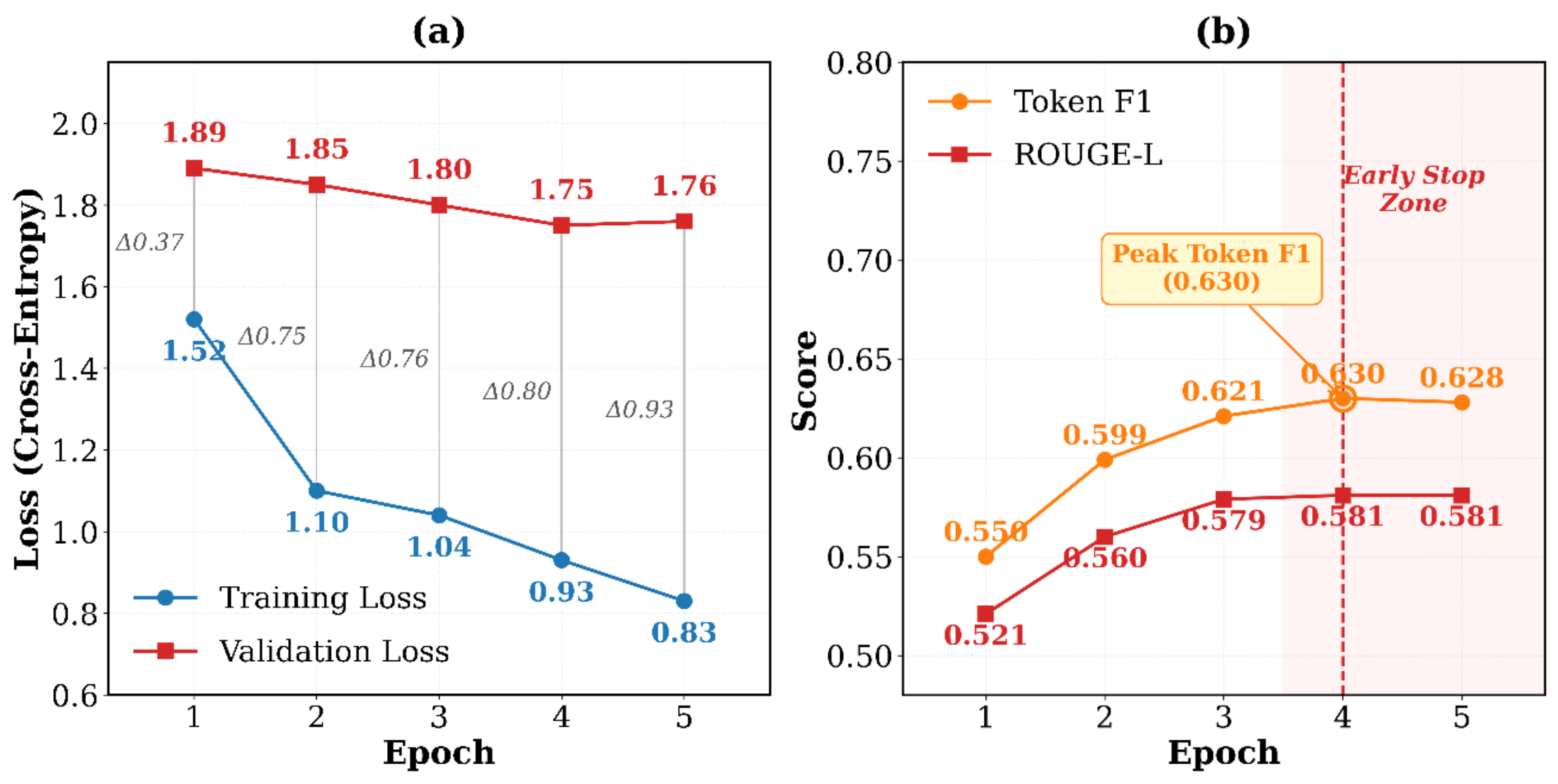

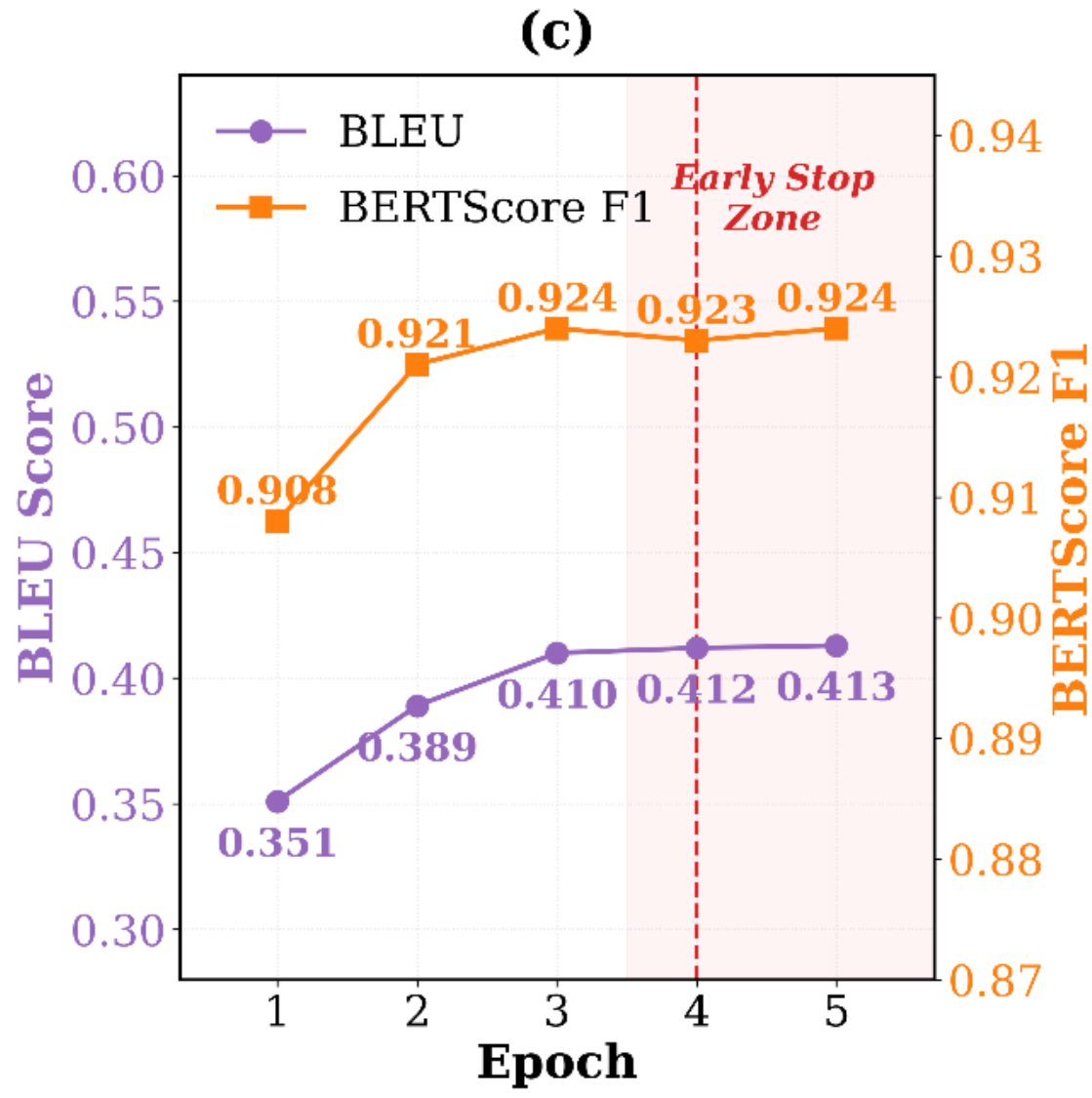

**Figure 3. Complete training dynamics of Llama 3.1 8B model**
**(a)** Loss convergence with training/validation gap indicating healthy generalisation; **(b)** Token level metrics (F1, ROUGE-L) peaking at epoch 4, triggering early stopping; **(c)** Semantic metrics (BLEU, BERTScore) stabilising with minor noise.

### 2.2 Retrieval augmentation transforms factual grounding across all three architectures

Integrating the three fine-tuned models with a hybrid retrieval pipeline, combining SPECTER2 dense embeddings over 68,039 paper chunks, BM25 lexical retrieval, reciprocal rank fusion, cross-encoder reranking, and maximal marginal relevance diversification, produced large, consistent improvements in answer quality across all architectures. For Llama-3.1-8B, Token F1 increased from 0.265 ± 0.080 to 0.686 ± 0.172 (+159%), ROUGE-L from 0.215 to 0.698 (+225%), and BLEU from 0.054 to 0.480 (+789%). For Qwen-2.5-7B, Token F1 rose from 0.242 ± 0.080 to 0.587 ± 0.191 (+143%). Mistral-7B showed the largest absolute gains as improvements were evident in Token F1 from 0.250 ± 0.081 to 0.737 ± 0.208 (+195%), ROUGE-L from 0.184 to 0.740, and BERTScore F1 from 0.866 ± 0.020 to 0.951 ± 0.035 (+10%), making it the top-performing configuration across all lexical and semantic metrics. Across all three models, RAG-induced Token F1 gains ranged from 143% to 194%, lifting baselines uniformly from the 0.24–0.25 range to 0.59–0.74. (Figure 4).

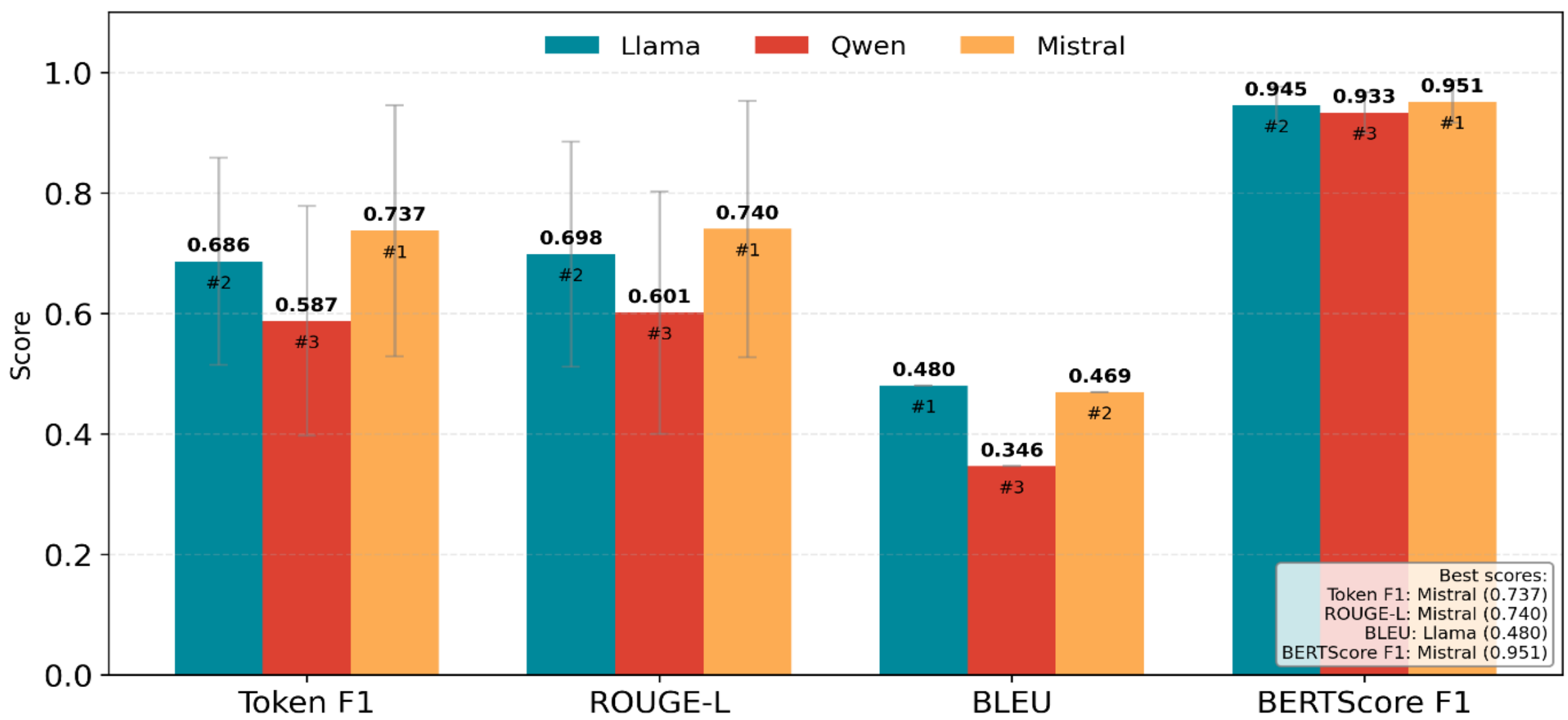

**Figure 4. Comprehensive three-model comparison across four complementary metrics with RAG.** Mistral shows strong results across Token F1 (0.737), ROUGE-L (0.740), and BERTScore (0.951). Llama leads in BLEU with 0.480 and has the lowest variance. Qwen performs adequately on all metrics but falls short on Token F1 and ROUGE-L.

To better understand these gains, we quantified the gap between semantic similarity and lexical grounding as BERTScore F1 minus Token F1. This semantic–lexical grounding gap provides a practical way to identify responses that sound plausible but remain weakly anchored to domain-specific factual content. Without retrieval, all three models showed large gaps of approximately 0.60: 0.601 for Llama, 0.621 for Qwen, and 0.616 for Mistral. This indicates that the models could produce semantically relevant answers but often missed precise terminology, factual details, alloy designations, corrosion mechanisms, and the experimental conditions expected in expert responses. Retrieval augmentation substantially reduced this gap, to 0.259 for Llama, 0.346 for Qwen, and 0.214 for Mistral. The reduction was driven mainly by large improvements in Token F1, with only modest gains in BERTScore F1 of approximately 8-10%. This suggests that RAG improves factual grounding rather than simply preserving fluent or semantically plausible output. Mistral showed the smallest residual gap, indicating the closest alignment between semantic plausibility and token-level factual accuracy (Figure 5, Table S5).

Paired bootstrap testing with 10,000 resamples did not reach statistical significance at the current test-set size of 140 questions, with $p$-values of approximately 0.50. However, the observed effect sizes were large across all three architectures, with Cohen's $d$ values ranging from 2.1 to 3.1, indicating practically meaningful improvements. The discrepancy between large effect sizes and non-significant $p$-values reflects high inter-question variance in the test set, which inflates standard errors despite consistent mean improvements, a pattern characteristic of small-sample evaluations on heterogeneous question sets, and one that motivates larger-scale validation in future work. Latency profiling showed that retrieval added only 2.4-4.6 s per query, whereas generation remained the main computational cost at 18.2-24.0 s across models. End-to-end response times were 20.6 s for Qwen, 25.4 s for Mistral, and 28.6 s for Llama, revealing a clear speed-accuracy trade-off: Qwen was the fastest, Mistral achieved the highest accuracy, and Llama produced the most stable outputs.

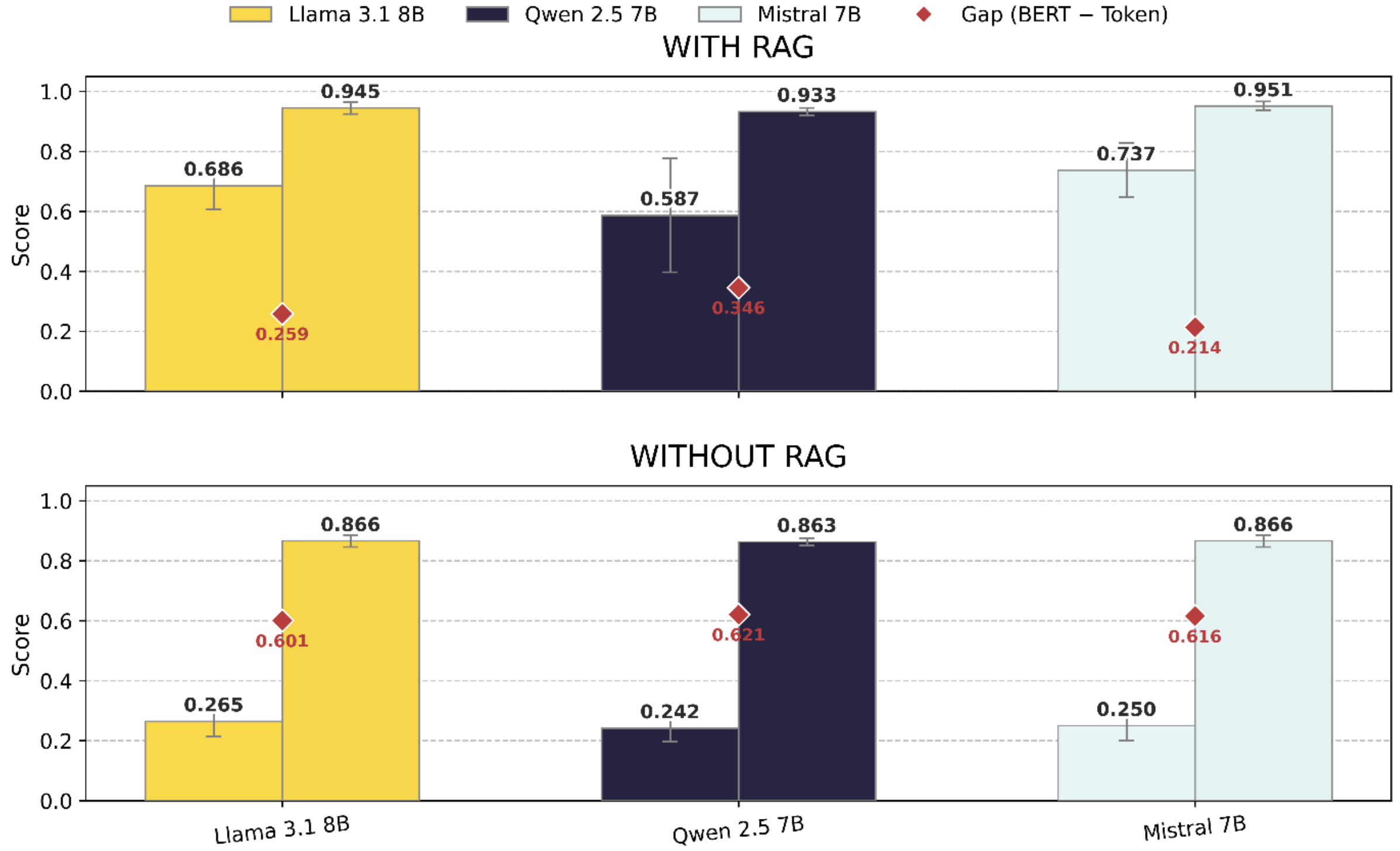

**Figure 5. Retrieval augmentation narrows the gap between semantic plausibility and lexical grounding.** Semantic-lexical divergence for the three fine-tuned models with and without RAG. For each model, the left bar shows Token F1, reflecting lexical overlap with the reference answers, and the right bar shows BERTScore F1, reflecting semantic similarity. The red diamond marks the grounding gap (BERTScore F1 - Token F1), which serves as a practical indicator of how far a response may appear semantically plausible while remaining weakly grounded in the expected technical content. Without retrieval (bottom), all three models show large grounding gaps of approximately 0.60 (Llama: 0.601, Qwen: 0.621, Mistral: 0.616), indicating that their answers often sound plausible but lack strong factual anchoring. With RAG (top), these gaps shrink to 0.259 for Llama, 0.346 for Qwen, and 0.214 for Mistral, driven mainly by large gains in Token F1 alongside smaller increases in BERTScore F1. This pattern suggests that retrieval improves factual grounding rather than simply maintaining semantic fluency.

### 2.3 Automated and expert evaluation reveals reliable grounding and model-specific trade-offs

Independent evaluation using the RAGAS framework, with GPT-4o-mini as a fixed LLM judge across all three models, broadly supported the token-level and semantic results. All models achieved high faithfulness scores, with 0.964 for Llama, 0.961 for Mistral, and 0.935 for Qwen, together with consistently high context recall values of 0.988–0.993. This indicates that the hybrid retrieval pipeline reliably retrieved relevant evidence and grounded the generated answers across all downstream models, an important requirement for safety-critical applications. Context precision was also strong and closely matched across models, ranging from 0.910 to 0.918. Clearer model-specific differences appeared in answer relevancy and answer correctness. Llama achieved the highest answer relevancy, scoring 0.748 compared with 0.695 for Qwen and 0.672 for Mistral, suggesting that its responses were more focused on the specific question. Mistral achieved the highest answer correctness, with a score of 0.920 compared with 0.847 for Llama and 0.841 for Qwen, consistent with its stronger Token F1 performance. Because context precision and recall were nearly uniform across the three models, these differences are more likely to reflect generation behaviour than retrieval quality (Figure 6).

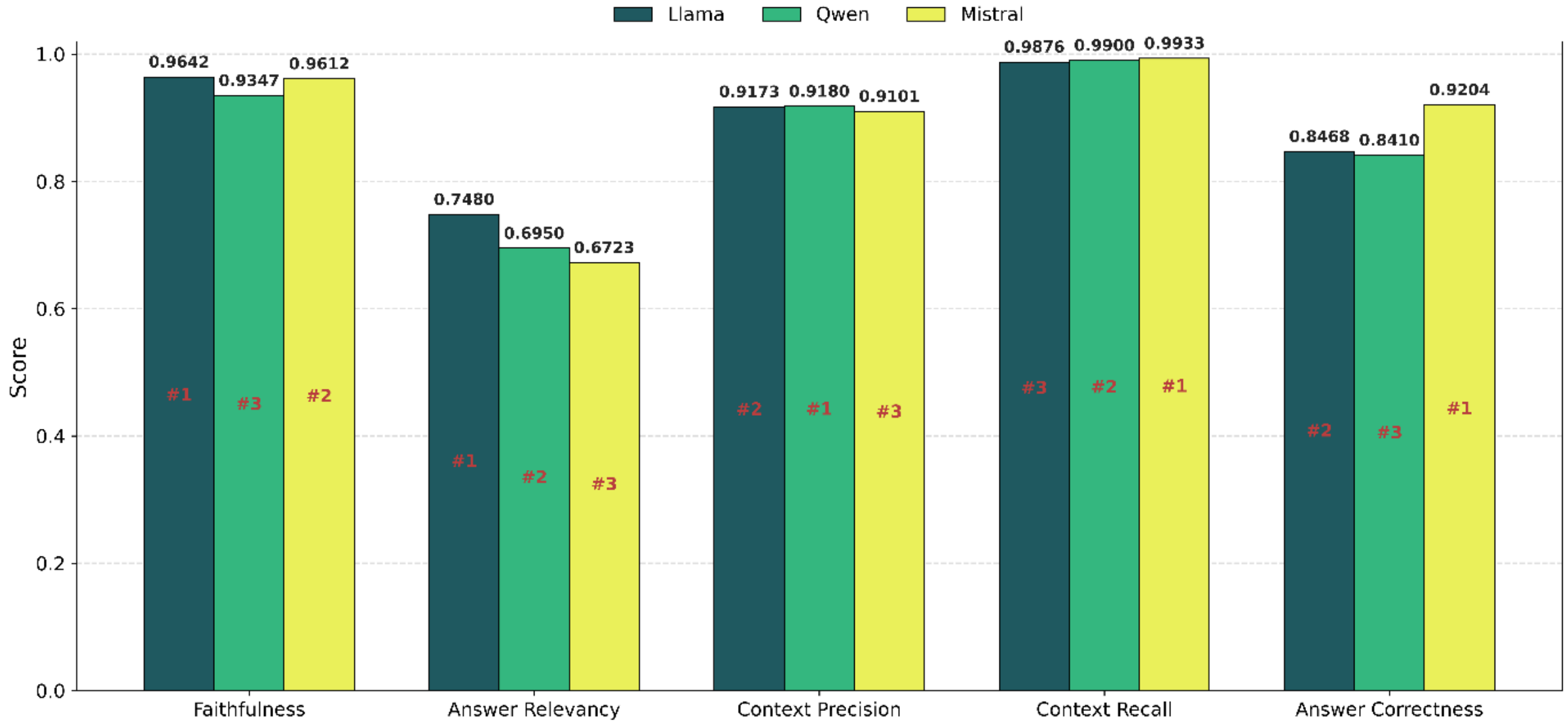

**Figure 6. RAGAS metrics comparison**

RAGAS evaluation across five complementary metrics. Llama leads on Answer Relevancy (0.7480). Qwen leads on Context Precision (0.9180). Mistral achieves the highest Answer Correctness (0.9204) and Context Recall (0.9933). All models show high Faithfulness (~0.96).

A blinded multi-expert evaluation provided a complementary assessment of model performance across 20 technically demanding questions spanning magnesium corrosion, corrosion-fatigue, electrochemical modelling and biomedical implant applications. With model identities concealed throughout, the evaluation broadly supported the automated findings while adding a practical dimension that automated metrics cannot fully capture: whether the responses are useful for research and engineering decision-making. Llama achieved the highest overall mean expert score, with 3.93 out of 5.00, followed closely by Mistral at 3.88 and Qwen at 3.80. Technical accuracy was the strongest dimension across all models, with scores of 4.11 for Llama, 4.09 for Mistral and 3.96 for Qwen, indicating that retrieval grounding generally supported scientifically sound generation. Content completeness was the most consistent limitation, with scores of 3.84 for Llama, 3.78 for Mistral and 3.75 for Qwen. Expert comments noted occasional omissions of relevant mechanisms, particularly in biomedical implant design questions. Practical relevance followed a similar pattern, with Llama scoring 3.85, Mistral 3.78 and Qwen 3.70, suggesting that the responses were generally understandable but sometimes lacked explicit engineering-oriented interpretation (Figure 7, Table S6). The relationship between RAGAS and human evaluation is also informative. Llama's leading overall expert score is consistent with its highest RAGAS answer relevancy, whereas Mistral's competitive expert-rated technical accuracy, comparable to Llama's, aligns with its superior RAGAS answer correctness. At the same time, the persistent gap between technical accuracy and completeness shows that factual correctness alone does not guarantee complete or actionable answers. For deployment in research, engineering design or biomedical translation, responses must be not only correct, but also complete, contextualised and practically interpretable.

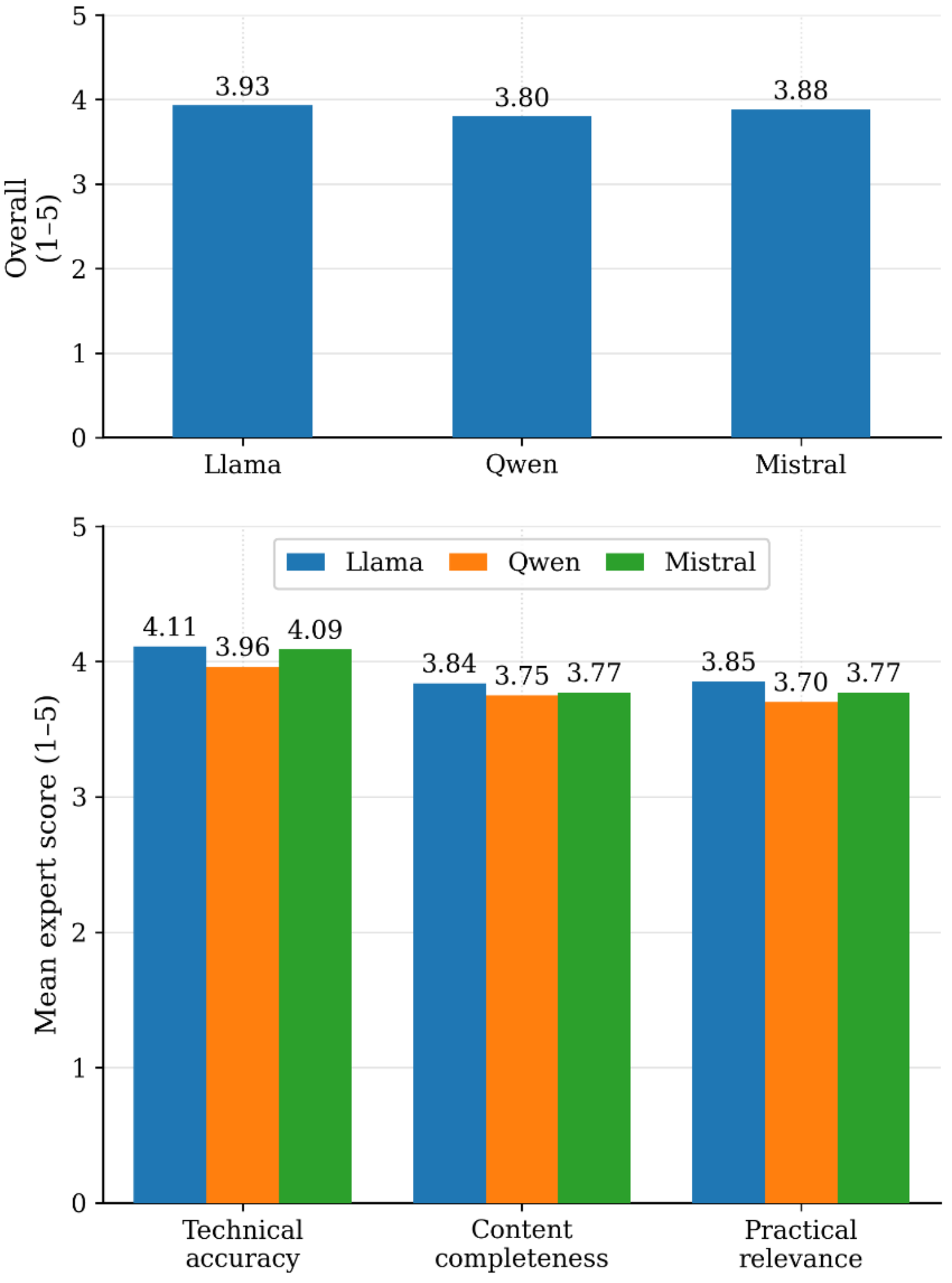

**Figure 7. Multi-expert human evaluation of RAG outputs across three fine-tuned models.**
Top: overall mean expert scores on a 1–5 Likert scale, with Llama leading at 3.93, followed by Mistral at 3.88 and Qwen at 3.80. Bottom: mean scores for the three evaluation dimensions assessed independently, technical accuracy (Llama: 4.11, Mistral: 4.09, Qwen: 3.96), content completeness (Llama: 3.84, Mistral: 3.77, Qwen: 3.75), and practical relevance (Llama: 3.85, Mistral: 3.77, Qwen: 3.70). Model identities were concealed throughout the evaluation. Technical accuracy was the strongest dimension across all three models, while content completeness and practical relevance were the most consistent limitations, with a gap between accuracy and completeness evident across all models.

### 2.4 Out-of-distribution validation confirms trend-level generalisation with residual weaknesses in fine-grained mechanistic fidelity

To assess generalisation beyond the training distribution, to our knowledge, the first such test for an LLM-based corrosion question-answer (Q-A) system, the three RAG models were evaluated on questions derived from three recently published magnesium alloy corrosion papers, entirely excluded from both training and indexing. Six mechanistically non-trivial examples were evaluated blind, spanning the contrasting effects of bovine serum albumin on stress corrosion cracking versus potentiodynamic polarisation in Hanks' solution;[21] the AZ-series corrosion paradox, in which β-phase morphology and eutectic α interactions produce a non-intuitive corrosion resistance ranking; [22] and friction stir processing effects on ZE52 alloy corrosion, where the finest-grained condition is not the most corrosion resistant.[23] Each example was scored across environmental classification, trend ranking, and mechanistic explanation (2 points each, 36 points total across the six examples). Full descriptions of the six examples, including the source-paper questions, retrieved evidence, and scoring rubric, are provided in the Supplementary document.

Llama achieved the highest cumulative score (24/36), followed by Qwen (20/36) and Mistral (10/36). Llama scored 5 out of 6 on four examples and was strongest in mechanistic explanation. Its responses closely followed the reasoning in the source papers, particularly in relation to precipitate homogenisation, solute re-solution, and alloy–oxide film chemistry. Qwen performed well in identifying corrosion trends and ranking corrosion resistance but was less reliable in explaining the underlying mechanisms. In some cases, it predicted the correct trend but attributed it to the wrong physical cause. Mistral showed the most variable performance: it could score well on individual sub-dimensions but often failed to combine condition classification, trend ranking, and mechanistic explanation into a coherent answer. A consistent pattern across all models was stronger performance on trend-level predictions than on mechanistic explanation, reflecting a known limitation of retrieval-augmented systems in reproducing fine-grained causal reasoning tied to individual experimental datasets (Figure 8). Notably, Mistral's pattern here inverts its earlier position on in-distribution RAGAS answer correctness, where it led with the highest score (0.920). This divergence shows that in-distribution factual correctness does not predict out-of-distribution mechanistic reasoning, and motivates the structural evaluation developed in the next section.

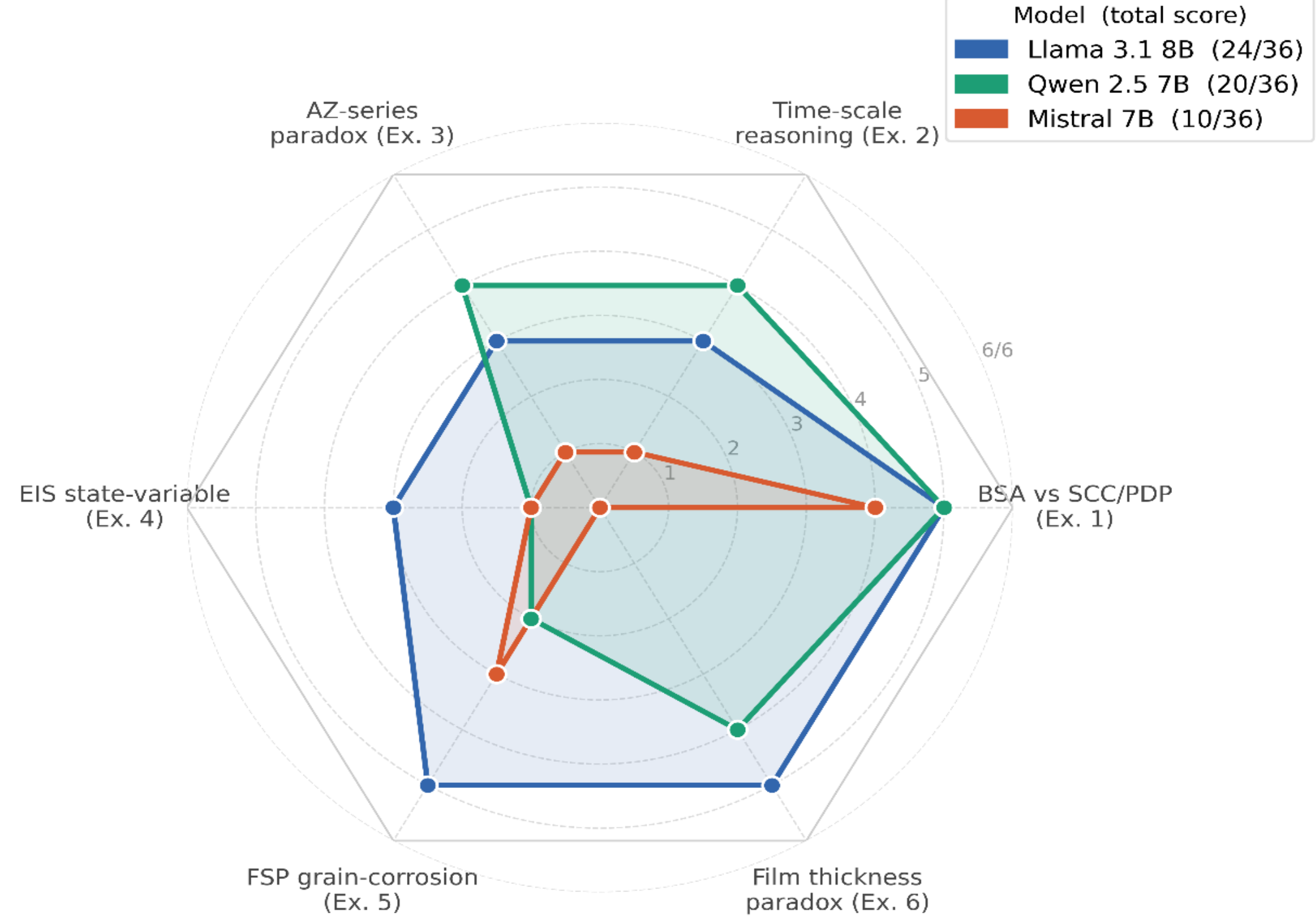

**Figure 8. Blind out-of-distribution evaluation of the three fine-tuned RAG models on six mechanistically non-trivial examples from unseen, recently published Mg alloy corrosion literature.**
Each axis represents one example: BSA vs SCC/PDP (Ex. 1), time-scale reasoning (Ex. 2), AZ-series corrosion paradox (Ex. 3), EIS state-variable interpretation (Ex. 4), FSP grain–corrosion decoupling (Ex. 5), and film thickness paradox (Ex. 6). Each example was scored 0–6 across three sub-dimensions: environmental/condition classification, trend ranking, and mechanistic explanation (2 points each). Total scores out of 36 are shown in the legend. Llama 3.1 8B (blue, 24/36) shows the most consistent and broadly extended profile, scoring 5/6 on four examples. Qwen 2.5 7B (green, 20/36) performs competitively on trend-based examples but contracts on EIS state-variable and FSP grain–corrosion tasks. Mistral 7B (orange, 10/36) shows a markedly reduced and irregular profile, reflecting difficulty integrating the three sub-dimensions into coherent responses. Detailed descriptions of each example, including the underlying source paper, question structure, and scoring criteria, are provided in the Supplementary document.

Independent experimental validation on an entirely new electrochemical dataset, open circuit potential transients and electrochemical impedance spectroscopy recorded at four timepoints for a Mg–Ca binary alloy immersed in Hanks' balanced salt solution at 37°C, not disclosed to any model at any stage, tested the same generalisation boundary with real measured data. Three diagnostic queries targeted the most discriminating features: the non-monotonic low-frequency impedance trajectory ($|Z|_{0.01}$ Hz rising from 3,951 Ω at 30 min to a peak of 6,912 Ω at day 1, then declining to 3,310 Ω at day 7, consistent with transient film formation followed by chloride-assisted degradation); the OCP evolution across immersion time; and the interpretation of the stable 7-day OCP (−1.834 V) as passivation versus kinetically stable active corrosion (Detailed descriptions of each queries are provided in the supplementary document). Llama achieved the highest aggregate score (7/9), followed by Mistral (6/9) and Qwen (5/9). Llama correctly identified the non-monotonic impedance turning point at day 1 and clearly rejected the misinterpretation of a stable OCP as passivation, correctly integrating multiple electrochemical signals. Qwen performed best on sequential OCP reasoning but misplaced the impedance turning point. Mistral captured the broader narrative of film formation and degradation but was less reliable in identifying dataset-specific turning points or subtle mechanistic distinctions. Across all models, performance declined when the task required separating kinetic stabilisation from true passivation, two electrochemical states that can appear similar but arise from different mechanisms. This mirrors the pattern observed in the blind literature validation: the models can recover broad trends reliably but still struggle with fine-grained discrimination at dataset-specific inflection points (Figure 9, Tables S7–S8).

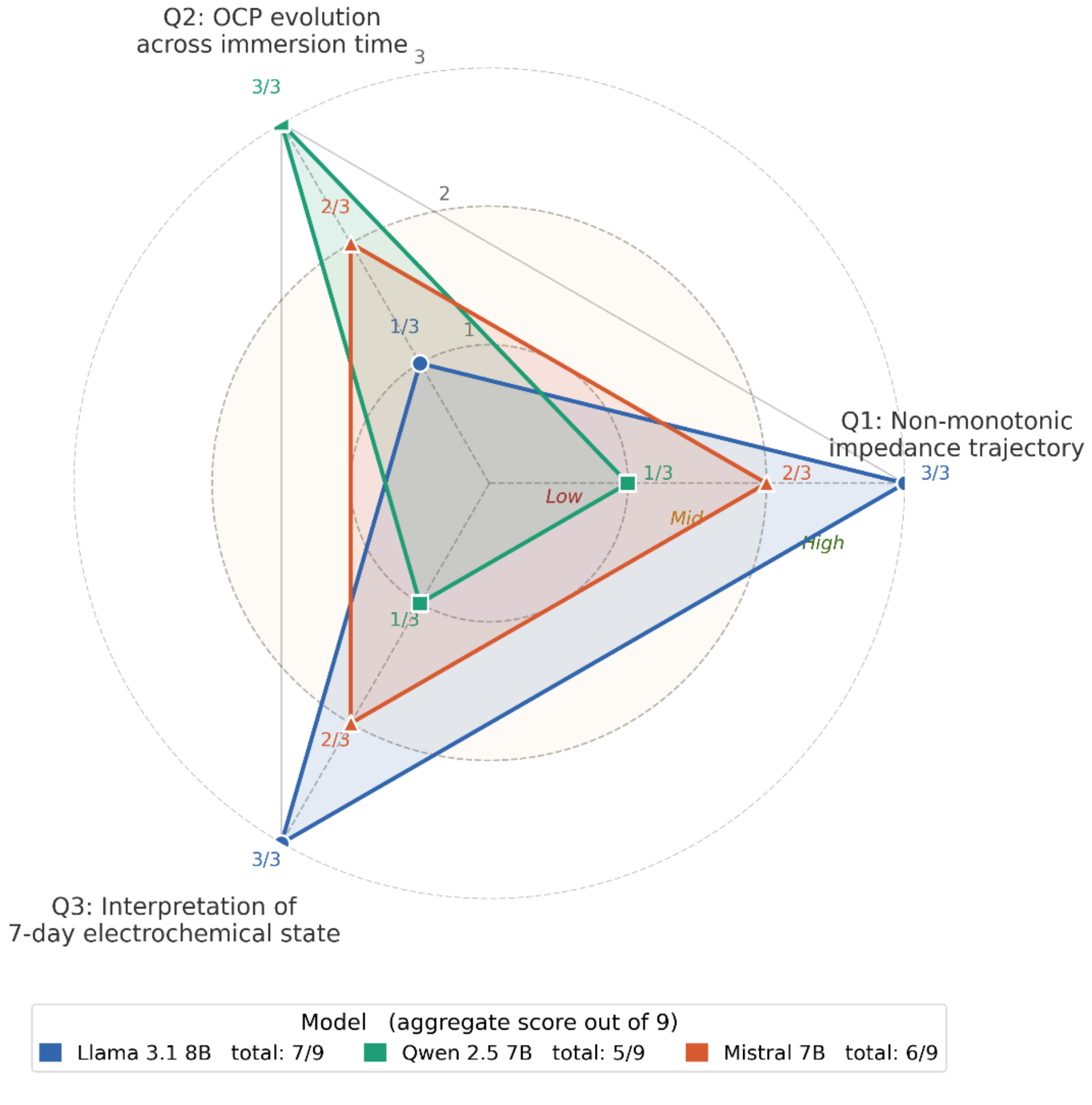

**Figure 9. Per-query reasoning profiles of the three fine-tuned RAG models on experimental electrochemical validation queries for a Mg–Ca binary alloy immersed in HBSS at 37 °C.** Each axis represents one diagnostic query: Q1 (non-monotonic impedance trajectory), Q2 (OCP evolution across immersion time), and Q3 (interpretation of the 7-day electrochemical state). Scores per query are annotated at each data point (maximum 3/3). Background shading indicates performance zones: green (high, 3/3), amber (mid, 2/3), and red (low, 1/3). Aggregate scores out of 9 are shown in the legend. Llama leads on the two interpretation-intensive tasks (Q1 and Q3), Qwen leads on OCP sequential reasoning (Q2), and Mistral occupies a consistent but less precise intermediate position across all three queries. Detailed descriptions of each query, including the underlying experimental data and scoring rubric, are provided in the Supplementary document.

## 2.5 Reason Map detects reasoning errors missed by claim-level evaluation

The results above show that high factuality scores do not necessarily mean that a generated answer is mechanistically sound. A model may achieve high RAGAS faithfulness and strong expert ratings while still linking evidence to conclusions in the wrong causal order. To capture these failures, we developed Reason Map, a three-phase proposition-graph framework that evaluates not only *what* a model says but also *how* it connects evidence into a reasoning chain. In Phase 1, the generated answer is broken down into atomic, falsifiable propositions. The relationships between proposition pairs are then classified into typed inference categories: *supports*, *implies*, *mechanistic-cause*, *contradicts*, *exception-to*, and *qualifies*. Each proposition is also checked against the retrieved evidence using a DeBERTa-v3-large NLI model,[24] producing proposition-level measures of evidence coverage and hallucination risk. In Phase 2, an independent evidence graph is constructed directly from the retrieved chunks, without reference to the generated answer, and aligned against the answer graph through proposition-level embedding similarity and NLI-based matching, enabling edge-level classification of each reasoning step as valid, unsupported-leap, or causal-reversal. In Phase 3, targeted self-correction is triggered when failures exceed predefined thresholds, with query decomposition, retrieval retry, and selective answer revision applied (Figure 10).

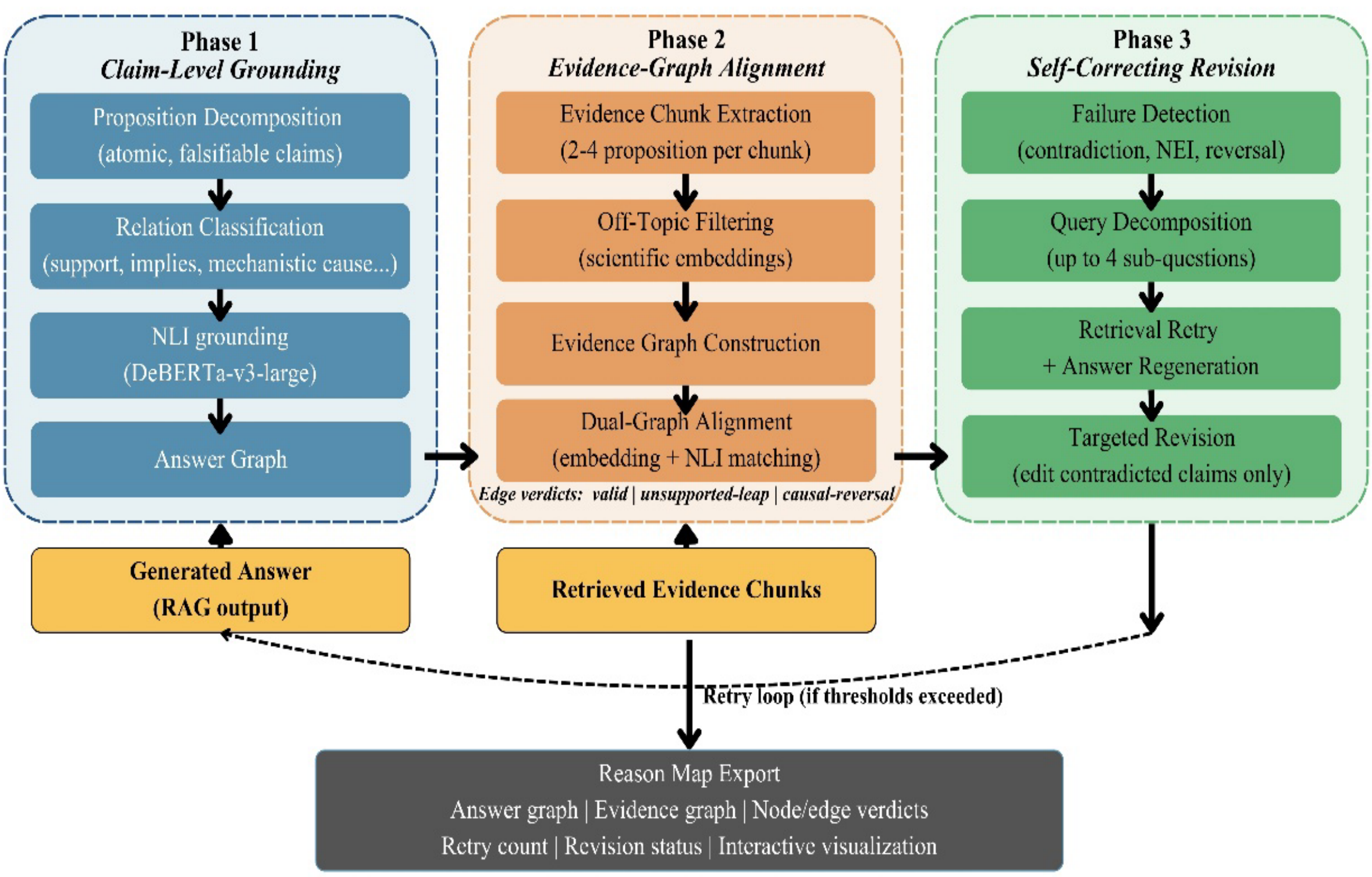

**Figure 10. Three-phase validation pipeline of Reason Map.** In Phase 1 (left), the generated RAG answer is decomposed into atomic propositions, pairwise inference relations between them are classified, and each proposition is checked against the retrieved evidence chunks using a DeBERTa-v3-large NLI model. The output of this phase is the answer graph. In Phase 2 (centre), an independent evidence graph is constructed directly from the retrieved chunks by extracting 2–4 atomic propositions per segment, filtering off-topic nodes with scientific-domain embeddings, and classifying inter-evidence relations. The answer graph and evidence graph are then aligned through embedding similarity and NLI-based matching, producing edge-level verdicts of *valid*, *unsupported-leap*, or *causal-reversal*. In Phase 3 (right), validation failures trigger targeted self-correction. The original query is decomposed into up to four focused sub-questions, retrieval is repeated, and answer regeneration or targeted revision is applied. If predefined thresholds remain exceeded after revision, a retry loop returns to Phase 1. The final export records both graphs, all node- and edge-level verdicts, the retry count, and the revision status, allowing interactive human inspection of the full validation trace.

The rare-earth elements case study presented in Figure 1 illustrates this dissociation directly. The generated answer scored a RAGAS faithfulness of 0.964, with seven of nine proposition nodes individually supported by retrieved evidence (NLI entailment 0.919–0.998), and no nodes were directly contradicted. Dual-graph alignment nonetheless identified a causal reversal at edge P4→P5. The answer asserted that denser RE oxide surface films cause coherent growth along the MgO (002) plane. The independently constructed evidence graph supports the reverse: coherent RE oxide nucleation along the (002) plane produces the observed film density, not the other way around[25]. Although P4 and P5 each passed NLI verification individually (entailment 0.995 and 0.919), the directional error was detectable only at the level of the typed inference edge. The same response also contained six unsupported inferential leaps (P1→P2, P2→P3, P5→P6, P6→P7, P7→P8, P9→P8) and two NEI (not-enough-information) nodes (P8, P9), including an unanchored Pilling–Bedworth ratio (PBR) value of 1.1653 for the $RE_2O_3$ layer, a hallucination risk that was not captured by retrieval-quality metrics. The value 1.1653, reported with four-decimal precision, did not appear in any retrieved chunk. A PBR in this range is plausible for rare-earth oxide layers (1 < PBR < 2 indicates a protective film[26]), but the specific numerical value cannot be traced to the evidence base, identifying it as a hallucinated quantitative claim.

A second case study examined the model's reasoning on hydrogen-assisted stress corrosion cracking, a failure mode mechanistically distinct from the surface-film phenomena explored in the rare-earth case. In response to the query *"What is the role of hydrogen in Mg alloy SCC mechanism?"*, the system generated a structured answer spanning three alloy systems: hydrogen embrittlement and crack propagation in WE43, coating-mediated suppression of hydrogen evolution in AZ31 (Sr-HA), and hydrogen accumulation at pit bases and crack tips in AZ91D. Three of five propositions were NLI-supported with entailment scores exceeding 0.99, none were directly contradicted, and EG node coverage reached 100%, again, an apparently well-grounded answer by conventional standards (Figure 11).

Reason Map nonetheless identified two causal reversals at edges P5→P1 and P4→P1, in which the answer inverted the direction of causation linking hydrogen accumulation and embrittlement to the general SCC mechanism. It also flagged two unsupported inferential leaps at P3→P2 and P3→P5, where the answer drew connections between coating-related findings (Sr-HA on AZ31) and alloy-specific mechanisms (WE43, AZ91D) without evidence-graph support (Figures 11 and 12). Phase 3 self-correction reduced the factual error rate to zero but did not eliminate the reversed dependencies. This indicates that post-hoc correction of structural errors is harder than preventing them at generation time, motivating future integration of Reason Map signals as training-time alignment objectives rather than as audit-only tools. Across both case studies, Reason Map demonstrates that claim-level evidence support does not guarantee reasoning-level validity, and that this failure carries practical consequences. In the rare-earth elements case (Figure 1), the causal inversion determines whether a practitioner should target film nucleation mode or film density as the primary design variable, a distinction with direct engineering implications.

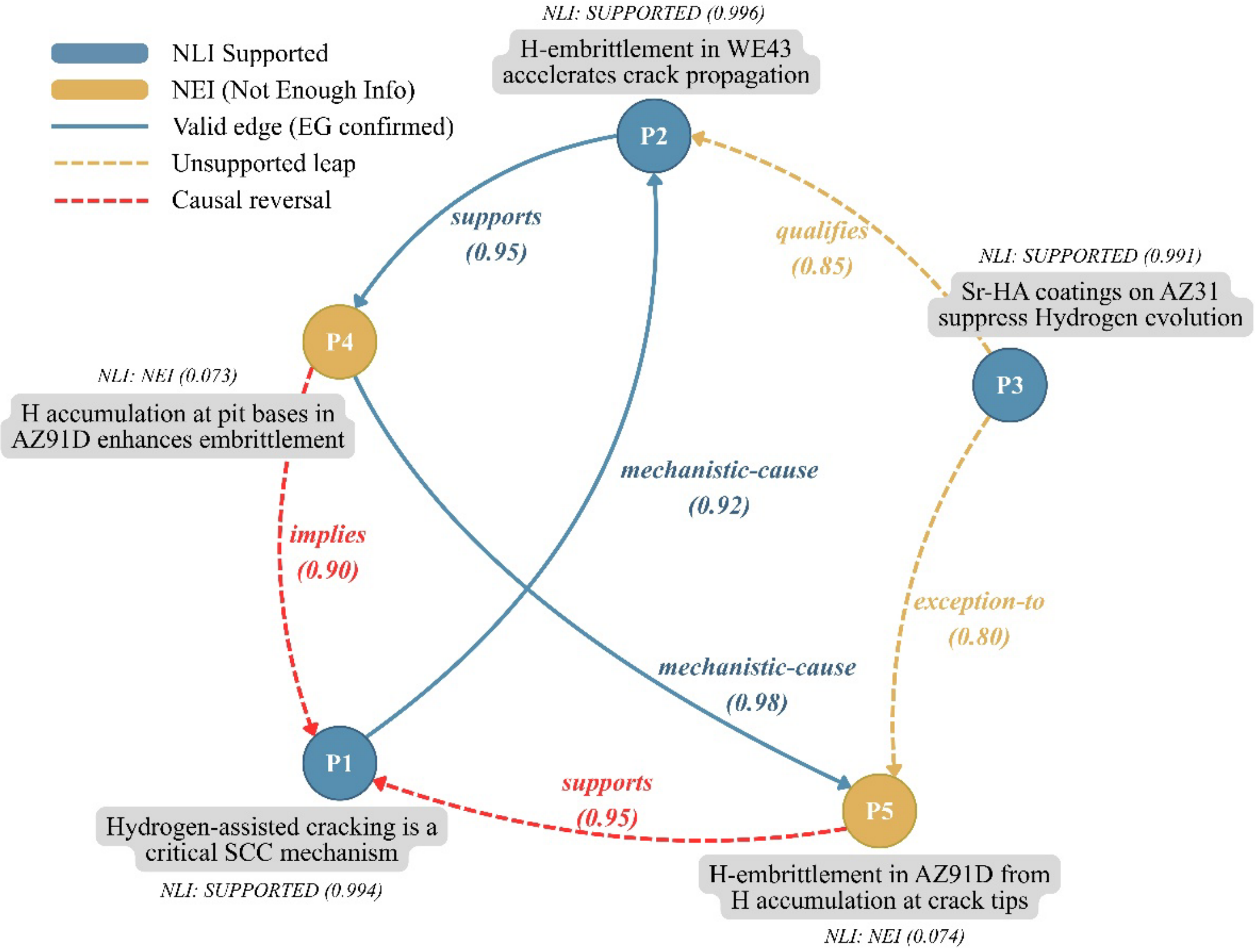

**Figure 11. Answer graph for the hydrogen-assisted SCC query.**
Each node represents an atomic proposition extracted from the generated answer, with node colour indicating the NLI verdict (blue: supported; amber: NEI). Propositions P1 (hydrogen-assisted cracking as a critical SCC mechanism; entailment 0.994), P2 (hydrogen embrittlement in WE43 accelerating crack propagation; 0.996), and P3 (Sr-HA coatings on AZ31 suppressing $H_2$ evolution; 0.991) are strongly anchored to retrieved evidence. P4 and P5, describing hydrogen accumulation and embrittlement in AZ91D, received NEI verdicts (entailment 0.073 and 0.074). Directed edges are coloured by evidence-graph validation: solid blue denotes valid edges confirmed by a forward path in the evidence graph (P1→P2, P2→P4, P4→P5); dashed amber denotes unsupported leaps with no evidence-graph path (P3→P2, P3→P5); and dash-dot red denotes causal reversals where only the reverse direction is supported (P5→P1, P4→P1). Edge labels show the classified relation type and confidence score.

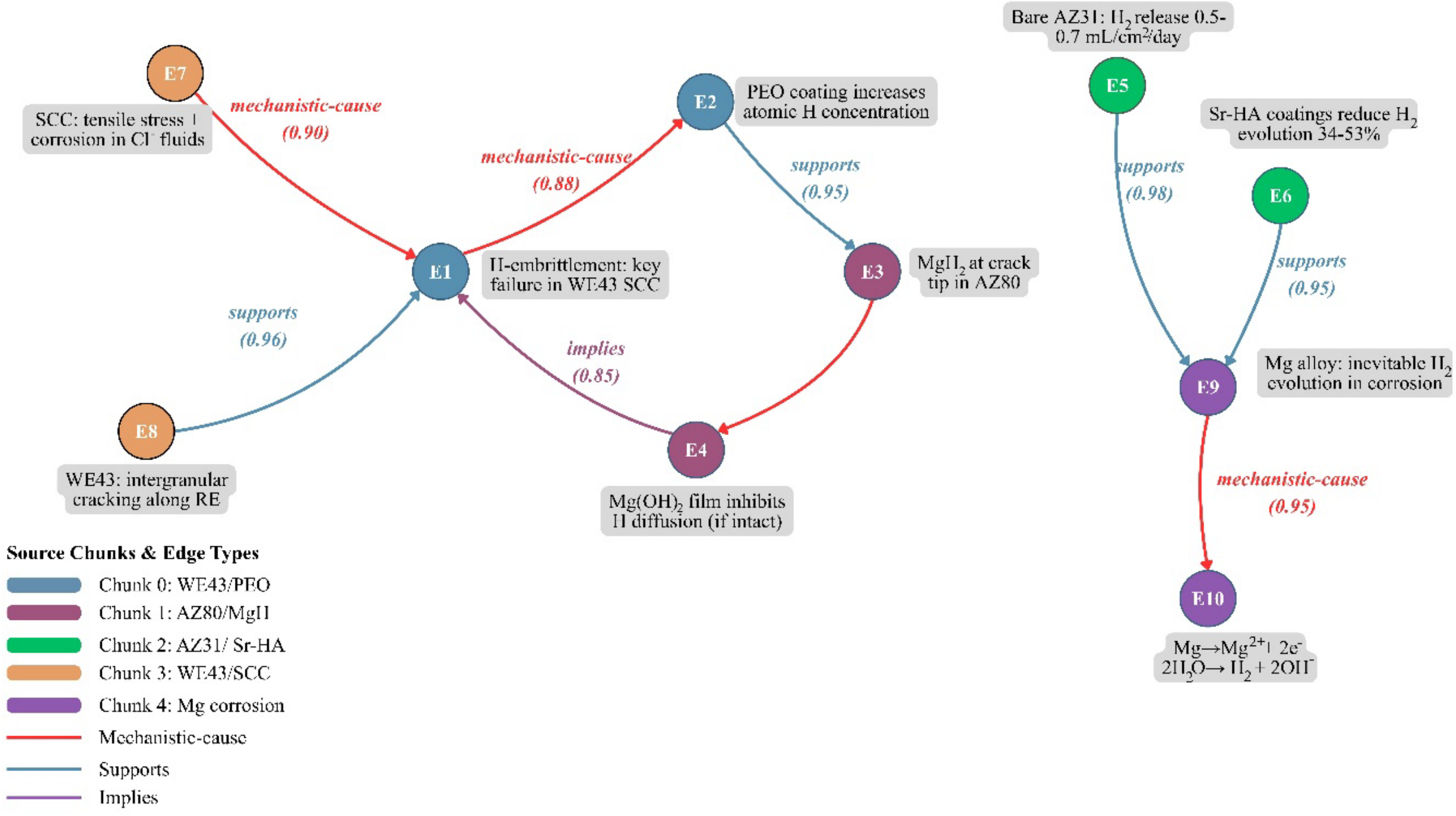

**Figure 12. Evidence graph constructed independently from the top five retrieved chunks.**
Each node (E1–E10) represents an atomic factual proposition extracted from the literature, with node colour indicating the source chunk: Chunk 0 (blue, WE43/PEO coating), Chunk 1 (mauve, AZ80/$MgH_2$), Chunk 2 (green, AZ31/Sr-HA coatings), Chunk 3 (orange, WE43/SCC mechanisms), and Chunk 4 (purple, general Mg corrosion electrochemistry). Directed edges represent evidence-supported inference relations: red for *mechanistic-cause*, blue for *supports*, and purple for *implies*, each annotated with a confidence score. The graph contains two principal sub-networks: a hydrogen-embrittlement cluster (E1–E4, E7–E8) centred on E1 as the convergence node for WE43 failure mechanisms, and a hydrogen-evolution pathway (E5→E6→E9→E10) linking experimental coating data to the fundamental anodic–cathodic corrosion reactions. This independently derived topology serves as the reference structure against which answer-graph edges are validated in Phase 2.

## 3. Discussion

The central finding of this study is that retrieval augmentation substantially improves the reliability of corrosion-focused language models, but does not by itself guarantee mechanistically sound reasoning. Across the three fine-tuned architectures, RAG increased Token F1 by 143–194%, while the leading system achieved high RAGAS faithfulness and context recall. These results show that the hybrid retrieval pipeline can consistently retrieve relevant evidence and anchor generated answers to it. At the same time, the semantic–lexical divergence analysis shows why retrieval is not a complete solution. Without RAG, all models produced large grounding gaps of approximately 0.60, indicating responses that were semantically plausible but weakly tied to domain-specific factual detail. Retrieval reduced these gaps to 0.21–0.35, closing much of the distance between fluency and accuracy, but not eliminating it. The complementary behaviour of the three models reinforces this interpretation (Figure 4). Mistral achieved the strongest factual accuracy, with the highest Token F1 and RAGAS answer correctness, while Llama showed the most balanced performance across relevance, completeness, and practical usefulness, supported by both automated and blinded expert evaluation. Qwen, in contrast, offered faster responses but weaker mechanistic consistency. These differences suggest that, within this parameter range, base architecture and pretraining alignment can matter as much as model size. More importantly, no single model dominated across all deployment priorities. The choice of model, therefore, depends on whether the intended use prioritises factual precision, response stability, interpretability, latency, or practical engineering relevance.

The most consequential finding, however, is not simply that retrieval improves performance, but that it exposes a deeper limitation in how generated scientific answers are currently evaluated. The Reason Map case studies show that an answer can achieve high RAGAS faithfulness, pass natural language inference claim verification, and still contain causal reversals or unsupported inferential links. In corrosion science, the order and direction of electrochemical, chemical, and transport processes can determine whether a practitioner targets film nucleation mode or film density, or whether a stable open-circuit potential is interpreted as passivation rather than active corrosion. These structural errors are not rhetorical; they shape the recommendations a practitioner acts on. Reason Map addresses this gap by shifting evaluation from isolated factual claims to the structure of the reasoning chain that connects them, providing a form of trust calibration that flat factuality metrics cannot offer. The Phase 3 self-correction results further clarify the challenge. Post-generation correction reduced factual errors, but some reversed causal dependencies remained. This suggests that factual content is easier to repair after generation than the structure of the reasoning itself. Models can be corrected to say fewer unsupported things, but it is harder to make them reorganise evidence into the right causal sequence once an incorrect reasoning structure has already been produced. This points to an important direction for future work: Reason Map signals, unsupported leaps, causal reversals, and NEI rates should be explored not only as post-hoc audit outputs but also as training-time alignment objectives.

Several limitations define the scope of the present study. The benchmark focuses on magnesium alloy corrosion, and broader validation across other alloy systems, degradation mechanisms, and exposure environments is needed before generality can be claimed. Blind external validation, to our knowledge, the first of its kind in LLM-guided corrosion research, was conducted on six carefully designed examples and does not yet constitute a large-scale out-of-distribution benchmark. The experimental electrochemical validation dataset was compact, with four timepoints, one alloy, and one electrolyte. Future validation should incorporate richer multimodal evidence, including microscopy, surface characterisation, long-term immersion data, and electrochemical datasets collected across a wider range of environmental conditions. The held-out test set of 140 questions was sufficient to demonstrate large practical effect sizes (Cohen's $d = 2.1–3.1$) but underpowered for definitive significance testing at conventional thresholds; larger evaluation sets will be needed to establish significance more robustly. Finally, Reason Map currently operates as a post-generation auditing layer, meaning that reasoning errors are detected and corrected after the answer is produced rather than avoided during training. Integrating its signals as training-time alignment objectives would address this limitation.

These limitations also define a clear path forward. Expanding the corpus and retrieval system into a broader multimodal corrosion intelligence platform would allow textual literature, electrochemical measurements, microstructural characterisation, and imaging data to be interpreted together. A temporally updated benchmark built from newly published papers would provide a stronger test of whether models can generalise to genuinely unseen literature as the field evolves. Most importantly, integrating Reason Map outputs into the model development loop could move the framework from reactive auditing toward proactive reasoning alignment. This would directly address the residual structural errors that survive standard retrieval, factuality checking, and post-generation correction.

More broadly, the framework introduced here is designed to be transferable beyond magnesium corrosion. The combination of hybrid retrieval, open-weight model adaptation, automated and expert evaluation, and proposition-graph reasoning validation can be applied to other materials science domains where mechanistic explanation matters, including alloy design, degradation modelling, heterogeneous catalysis, and biomaterials engineering. The results show that reliable LLM deployment in safety-critical materials science requires more than retrieval quality and answer-level faithfulness. It also requires domain-specific grounding, external and experimental validation, and explicit evaluation of reasoning structure. The RAG pipeline and Reason Map framework developed here provide a practical and reproducible foundation for more transparent, evidence-grounded, and scientifically defensible use of language models in corrosion research and related engineering fields.

## 4. Methods

### 4.1 Dataset construction and curation

A domain-specific corpus of 840 peer-reviewed open-access papers on magnesium alloy corrosion was assembled from publicly available sources, covering corrosion mechanisms, electrochemical behaviour, surface film formation, detection methods, and protective technologies. All articles were verified as available under open-access licences; no proprietary or paywalled materials were included (Figure S1, S2, S3). The corpus was processed using GROBID [27], an open-source machine-learning tool that converts PDF documents to TEI XML, to extract structured representations of titles, authors, abstracts, section headings, body text, and figure and table captions, subsequently post-processed into hierarchical JSON files (metadata). GROBID-based extraction succeeded for 99.9% of documents; the remaining 0.1% were processed using PyMuPDF [28] as a plain-text fallback. No documents failed extraction.

Question–answer (Q-A) pairs were generated from the structured text using a prompt-based LLM pipeline. Each generated pair was subjected to manual expert review covering factual accuracy, terminological precision, completeness, and absence of duplication. Pairs that failed one or more criteria were either revised or removed. The final verified dataset contained 3,309 question–answer pairs in JSONL format. Each record included a unique identifier, question, answer, structured metadata covering alloy composition, environmental conditions, corrosion metrics, and task annotations, and source attribution (Figure S4). To avoid information leakage between splits, we partitioned the data at the document level, so that all question–answer pairs derived from the same paper remained in the same subset. The final partitions contained 2,604 training pairs, 350 validation pairs, and 355 test pairs.

**4.2 Model selection and parameter-efficient fine-tuning**

Three open-weight instruction-tuned language models were selected for fine-tuning: Llama-3.1-8B-Instruct (Meta AI, 8.03B parameters),[29] Qwen-2.5-7B-Instruct (Alibaba Cloud, 7.66B parameters, 32,768-token context window, pretrained on 18 trillion tokens),[30] and Mistral-7B-Instruct-v0.3 (Mistral AI, 7.29B parameters, 32,768-token context window).[31] Selection criteria were: instruction-following capability, computational efficiency in the 7–8B parameter range suitable for single-GPU training, open-source availability with permissive licensing, and proven performance on scientific text understanding. All three models were fine-tuned under identical configurations to ensure that observed performance differences reflect architectural and pretraining characteristics rather than tuning variability.

Fine-tuning employed supervised learning on the curated Q-A pairs using Low-Rank Adaptation (LoRA),[32] a parameter-efficient strategy that introduces a low-rank decomposition $\mathbf{\Delta W} = \mathbf{A} \cdot \mathbf{B}$ to each weight matrix $\mathbf{W}$, where $\mathbf{A} \in \mathbb{R}^{(m \times r)}$ and $\mathbf{B} \in \mathbb{R}^{(r \times n)}$ with rank $r \ll \min(m, n)$, updating only a small fraction of parameters while keeping original weights frozen [33]. LoRA adapters (rank $r = 16$, scaling factor $\alpha = 32$, dropout 0.1) were applied to all key linear sub-modules, including query, key, value, and output projection matrices in self-attention and gated feed-forward network projections, yielding approximately 0.5% trainable parameters relative to the full model. To reduce GPU memory requirements, Quantized LoRA (QLoRA) [34] was applied, compressing base model weights to 4-bit NormalFloat4 precision while maintaining LoRA adapter weights and optimiser states in BF16.

All models were trained on a server equipped with dual NVIDIA Tesla V100 GPUs (32 GB VRAM each), 64 GB system RAM, and dual Intel Xeon Silver 4216 CPUs, using Hugging Face Transformers with Accelerate for distributed execution.[35] Inputs were capped at 4,096 tokens with dynamic padding; gradient checkpointing was enabled. Loss masking was applied so that only answer tokens contributed to the training objective, following each model's official chat template for tokenisation. Models were trained for five epochs using AdamW with a peak learning rate of $2 \times 10^{-4}$, linear decay, and a 6% warmup ratio. Effective batch sizes of 32 (Llama and Qwen) and 30 (Mistral) were achieved through gradient accumulation over micro-batches of 4 and 6 per device, respectively. Early stopping with patience of three epochs was monitored via Token F1 on the validation set. A fixed random seed of 42 was used throughout. Full training configurations are provided in Table S3.

Model performance was evaluated on a separate held-out test set of 140 Q-A pairs using four complementary metrics: Token F1 (lexical overlap),[36] ROUGE-L (longest common subsequence similarity),[37] BLEU (n-gram precision),[38] and BERTScore F1 (contextual semantic similarity using a pretrained language model encoder).[39] This multi-metric framework captures both surface-level terminological fidelity and deeper semantic alignment, both of which are critical in technical scientific domains that require precise domain-specific language. Performance comparisons of all three models, with and without RAG, across four metrics, along with latency profiles, are shown in the supplementary Figures S5-S10.

### 4.3 Knowledge base construction and hybrid retrieval pipeline

Followed by processing with GORBID, the body text was segmented into overlapping token-bounded chunks of 320 tokens with approximately 12% overlap (effective stride ≈ 282 tokens), yielding 68,039 paper chunks across the full corpus. Each chunk was encoded using SPECTER2,[40,41] a transformer-based model trained on scientific citation graphs, producing 768-dimensional dense embeddings optimised for scientific retrieval. In parallel, the 3,324 curated Q-A pairs were segmented into 3,920 answer-centric chunks of 240 tokens with approximately 12% overlap and encoded using BGE-M3,[42] producing 1,024-dimensional embeddings supporting dense, sparse, and multi-vector retrieval across inputs up to 8,192 tokens.

For both collections, parallel dense and lexical indices were constructed. SPECTER2 and BGE-M3 embeddings were L2-normalised and stored in FAISS indices[43] for approximate nearest-neighbour search; BM25 indices[44] were built over tokenised text for complementary sparse lexical retrieval. Chunk-level metadata including document identifier, section heading, and character offsets was preserved to enable transparent source attribution at inference time.
The query processing workflow proceeded as follows. Each user query was encoded in parallel by SPECTER2 (papers index) and BGE-M3 (Q-A index) to produce dense query embeddings, and simultaneously tokenised for lexical retrieval. Four retrieval modules ran in parallel, BM25 and FAISS-dense over both the paper and Q-A collections, each returning the top-60 candidate passages. The retrieval stage produced an initial pool of about 240 candidate passages, which were combined using Reciprocal Rank Fusion (RRF). This step merged the ranking signals from sparse and dense retrievers into a single shortlist. The top 100 fused candidates were then reranked using a BGE-M3 reranker, which scored each query-passage pair to refine relevance. After reranking, post-processing applied three constraints: a maximum of two passages per source document, at least one passage from the papers collection, and Maximal Marginal Relevance (MMR, $\lambda = 0.6$) to maintain diversity across retrieved perspectives.

### 4.4 RAG evaluation protocol

Each fine-tuned model was evaluated under two conditions on the held-out test set of 140 questions: a RAG condition in which the model received the top-10 retrieved passages (from a first-stage pool of 200 candidates, capped at 3,000 context tokens) produced by the hybrid retrieval pipeline, and a no-RAG baseline in which the identical model was prompted with the question alone. Both conditions used greedy decoding with a maximum of 128 generated tokens. Statistical significance of RAG versus no-RAG differences was assessed via paired bootstrap testing (10,000 resamples, two-sided, $\alpha = 0.05$). Effect sizes were computed as Cohen's d to quantify practical significance independently of sample size.

Automated RAG quality was assessed using the RAGAS framework,[43,44] which scores systems across five dimensions: faithfulness (answer consistency with retrieved context), answer relevancy (directness of response to the question), context precision (signal-to-noise ratio of retrieved passages), context recall (evidence coverage), and answer correctness (alignment with reference answer). Evaluation used GPT-4o-mini as a fixed LLM judge across all three models to ensure that observed RAGAS differences reflect RAG behaviour rather than variation in the evaluation rubric.

Human expert evaluation was conducted by domain experts who assessed RAG outputs from all three models across 20 technical questions spanning magnesium corrosion, corrosion-fatigue, electrochemical modelling, and biomedical implant applications. Model identities were concealed throughout. Each response was independently scored across three dimensions using a 5-point Likert scale: technical accuracy, content completeness, and practical relevance.

### 4.5 Out-of-distribution and experimental validation

Blind external validation was conducted on six questions derived from three papers on magnesium alloy corrosion published after the training and indexing cutoffs, and entirely absent from both the training and retrieval corpora. All questions were posed without providing any paper text to the models. Responses were scored against gold answers extracted directly from the source papers across three sub-dimensions: environmental and condition classification (A), trend ranking (B), and mechanistic explanation (C), each worth 2 points, yielding a maximum of 6 points per example and 36 total. The three source papers covered bovine serum albumin effects on stress corrosion cracking and potentiodynamic polarisation in Hanks' solution [21], the AZ-series corrosion resistance ranking paradox,[22] and friction stir processing effects on ZE52 alloy corrosion.[23]

Experimental validation was performed using electrochemical data from a Mg–Ca binary alloy immersed in Hanks' balanced salt solution (HBSS) at 37 °C. HBSS is a purely inorganic physiological electrolyte, with no protein or organic additives, and was chosen to focus the evaluation on model reasoning about inorganic film evolution. Open-circuit potential (OCP) transients were recorded over 3,600 s, and electrochemical impedance spectra were collected across 81 frequency points from 100 kHz to 0.01 Hz. Measurements were taken after 30 min, 1 day, 5 days, and 7 days of immersion. The full dataset (Table S7) was not disclosed to any model at any stage. Three validation queries were designed to test whether the models could interpret the non-monotonic low-frequency impedance trajectory, the evolution of OCP with immersion time, and the electrochemical state after 7 days. Each response was scored for directional accuracy, mechanistic fidelity, and quantitative consistency, with a maximum of 3 points per query and 9 points in total. (Queries and responses for both the external paper and the experimental validations are provided in Supplementary Notes S1 and S2)

### 4.6 Reason Map: proposition-graph reasoning validation

Reason Map is a three-phase post-generation auditing framework that validates the inferential structure of RAG-generated answers against an independently constructed evidence graph. In Phase 1, the generated answer was decomposed into atomic, falsifiable propositions using an LLM prompt that enforced single-claim granularity, domain-term preservation, and the retention of tension-bearing claims. Pairwise relations among propositions were classified into six typed inference categories: supports, implies, mechanistic-cause, contradicts, exception-to, and qualifies. Each proposition was grounded against the top retrieved evidence chunks using a DeBERTa-v3-large NLI model,[24] which assigned entailment, contradiction, or NEI (not enough information) verdicts, yielding the answer graph with proposition-level evidence coverage and hallucination risk measures.

In Phase 2, an independent evidence graph was constructed directly from the retrieved chunks without reference to the generated answer. For each of the top-ranked evidence segments, 2–4 atomic factual propositions were extracted, off-topic nodes were filtered using scientific domain embeddings, and inference relations among surviving evidence propositions were classified. The answer graph and evidence graph were aligned through proposition-level embedding similarity and NLI-based matching. Each answer-graph edge was then classified as: valid (a corresponding forward path exists in the evidence graph), unsupported-leap (no such path exists), or causal-reversal (only the reverse path is supported by the evidence topology). This dual-graph alignment distinguishes claim-level grounding from reasoning-level grounding.

In Phase 3, the framework performed targeted self-correction when validation failures exceeded predefined thresholds. The original query was decomposed into up to four focused sub-questions, retrieval was repeated over the expanded query set, and answer regeneration was triggered if contradicted claims, unsupported reasoning gaps, or causal reversals persisted. If only isolated contradicted claims remained, targeted answer revision edited only the affected statements or inverted causal directions while preserving the remainder of the response. The final Reason Map export recorded both graphs, all node- and edge-level verdicts, retry count, and revision status. An interactive web interface integrating the hybrid retrieval backend, three fine-tuned generation models, and the Reason Map engine was implemented in Gradio v4.0[45] (Figure S11).

## Conclusion

This work establishes that retrieval-augmented language models can support more reliable use in safety-critical corrosion phenomenon, particularly when their evaluation extends beyond the correctness of individual claims to the integrity of the reasoning that connects them. Underpinned by a sound example of the application of the model to magnesium alloy corrosion, the domain-specialised RAG pipeline introduced here improves unguided parametric generation by grounding responses in corrosion literature, technical evidence and experimental context. Its performance was validated across automated metrices, expert assessment, blind external literature tests and independent electrochemical experiments, demonstrating that retrieval can substantially improve the factual grounding of corrosion-specific answers.

The Reason Map framework complements this pipeline by auditing not only what a model states, but how it organises corrosion evidence into a mechanistic reasoning chain. This is critical in corrosion science, where the critical phenomena such as direction and sequence of electrochemical reactions, oxide-film formation, hydrogen evolution, transport processes and degradation pathways determine practical mitigation and design decisions. Reason Map exposed causal reversals, unsupported inferential leaps and misordered mechanistic explanations that flat factuality metrics could not detect, failures that, in a safety-critical context, can enable practitioner for more accurate recommendations.

Although developed and validated in corrosion domain, the combination of hybrid retrieval, multi-model fine-tuning, multi-paradigm evaluation and proposition-graph reasoning validation is domain-agnostic by design. It is therefore generic, and can be extended to other materials science contexts where trustworthy mechanistic explanation is essential, including alloy design, degradation modelling, heterogeneous catalysis and biomaterials engineering. More broadly, this work establishes a principle that extends beyond corrosion: in scientific domains where misplaced reasoning has real consequences, evaluating a language model's claims is necessary but not sufficient. The reasoning structure must be evaluated as well, and Reason Map provides a practical methodological foundation for doing so.

**Supplementary document**

**Table S1**: Reported language model applications in corrosion science

| Authors | Domain | Model / approach | Data size & type | Task(s) | Quantitative results (metrics) | limitations |
|---|---|---|---|---|---|---|
| Zhao & Birbilis, [1] 2023 | Corrosion protection | Word2Vec similarity search + BERT masked LM ("fill-mask" prompts) | 84M Scopus records → 5,990 corrosion papers → 1,812 filtered for training | Literature mining: candidate discovery and categorisation | Word2Vec: 54 relevant materials in top-1000 (83.3% benchmark-related rate). BERT: 30–85 relevant per prompt; combined BERT prompts: 161 relevant suggestions | Output depends on prompt/sentence structure; evaluation relies on benchmark categories and manual interpretation. |
| Wu et al. [2] 2025 | Corrosion | Qwen2.5-7B + LoRA SFT; CKD + RAG; embedding retrieval | 205 open-access corrosion papers; synthetic Q–A pairs with expert verification | Corrosion Q&A, risk identification, standards compliance, preventive guidance | Expert review: Risk ID: 88% "accurate"; Standards compliance: 75% "fully compliant"; Guidance: 90% "practical & effective". BLEU/ROUGE/BERTScore all improved vs baseline | Depends on knowledge-base completeness/timeliness; RAG adds runtime overhead; risk of confident errors without grounding |
| Busch et al. [3] 2025 | Corrosion / Mg | GPT-4o in-context learning with SMILES, names, experimental context ± molecular descriptors | 75 compounds (ZE41 Mg); labels = inhibition efficiency; baseline NN uses ~1,250 descriptors/sample (from prior dataset) | Regression: inhibitor efficiency prediction | Best GPT-4o (descriptors + pre-analysis): RMSE 50.5, MAE 43.4, r 85.8%. Context-only best (o1, SMILES only): MAE 51.6. Baseline MLP: RMSE 73.0, MAE 61.3, r 60.0% | Small dataset; output variance; imperfect determinism even with "seeded" outputs; potential hallucination |
| Jin et al. [4] 2025 | Corrosion / Mg | Comparative evaluation of LLM types and prompt designs; feature selection; few-shot prompting | 75 inhibitors (ZE41 Mg), six molecule categories; baseline ML uses selected descriptors | Regression: inhibitor efficiency prediction; prompt optimization. | DeepSeek-R1 (best configure): RMSE 52, MAE 41, r 84%, $R^2$ 0.66. Baseline ML: RMSE 71–79, MAE 61–63. Feature selection: "2–3 from 5" best; "5" alone gave RMSE 65, MAE 49 | Hallucination risk; performance sensitive to data quality/size; requires careful prompt engineering |
| Bongiorno et al. [5] 2026 | Corrosion (organic coatings | LLM-driven EIS interpretation for classification + parameter estimation | D1: 60 simulated spectra; D2: 71 exp. spectra; D3: 111 exp. spectra | Classification of EIS behaviour and parameter estimation | Classification accuracy: D1 96%, D2 88%, D3 97%. Parameter estimation errors comparable to trained NNs (MSE values in paper) | Prompt sensitivity; reduced accuracy for CPE parameters; limited circuit topologies; higher inference cost/latency |
| Bharath et al. [6] 2026 | Corrosion / Mg bioimplants | SciBERT fine-tuned for NER + LLaMA 3.1 relation extraction → knowledge graph; CoPILOT for protein folding simulation | 53 full-text corrosion articles; BSA protein structure (PDB ID: 4F5S) | Literature mining; knowledge graph construction; protein structure prediction | NER: F1 > 0.995, accuracy >99.8%. Predicted BSA structure: pLDDT = 86.53 (high confidence) | Limited training data (53 articles); LLaMA hallucinations; computational cost; no temporal/causal modeling |
| Sasidhar et al. [7] 2023 | Corrosion / alloys | NLP tokenisation + embedding + LSTM for text features + DNN for pitting potential | 769 records across five alloy classes; numerical + categorical + textual processing/test method descriptors | Regression: pitting potential prediction; optimisation | MAE ~150 mV; average $R^2 = 0.78 \pm 0.06$ (process-aware model) vs prior simple DNN $R^2 = 0.61 \pm 0.04$ | Outliers linked to missing processing-history text; optimisation can overshoot plausible compositions. |

| Authors | Domain | Model / approach | Data size & type | Task(s) | Quantitative results (metrics) | limitations |
|---|---|---|---|---|---|---|
| Yan et al., [8] 2024 | Mg phase diagram | GPT-4o/mini + prompt eng. + RAG (750k CALPHAD) + SFT (287k Q–A) | 750,000 CALPHAD points (Mg-Al-Zn); 287,523 for SFT | Phase diagram information acquisition / QA | Base models: 3–4/5. RAG: accuracy ↑. SFT: significant accuracy gain; direct answers; ε affects style not accuracy | RAG mismatches; catastrophic forgetting if narrow data; limited to Mg-Al-Zn; no thermodynamic understanding |
| Kumar et al. [9] 2024 | Mg alloys / Text mining | MagBERT (SciBERT-based continued pre-training) + MagNER (fine-tuned NER) | ~370K Mg abstracts (~67M words); ~140K words annotated NER (20 entities) | Masked language modeling; Named entity recognition; Mechanical property extraction | MLM: F1 0.74 (vs. 0.61–0.67 for general models). NER: Macro F1 0.93, weighted F1 0.96. Extracted ~520 clean composition-property entries from 16K abstracts | Training data bias; GPU requirements; struggles with complex context; heuristic post-processing needed; entity confusion possible |

**Table S2**. Architectural Comparison of Fine-Tuned Models

| Specification | Llama 3.1 8B | Qwen 2.5 7B | Mistral 7B v0.3 |
|---|---|---|---|
| **Total Parameters** | 8.07B | 7.66B | 7.29B |
| **Attention Mechanism** | Grouped Query Attention (GQA) | Multi-Head Attention (MHA) | Sliding Window Attention (SWA) |
| **Attention Heads** | 32 (8 KV heads) | 28 | 32 (8 KV heads) |
| **Transformer Layers** | 32 | 28 | 32 |
| **Hidden Dimension** | 4,096 | 4,096 | 4,096 |
| **FFN Dimension** | 14,336 | 11,008 | 14,336 |
| **Vocabulary Size** | 128,256 | 152,064 | 32,768 |
| **Max Context Length** | 8,192 | 32,768 | 32,768 (SWA: 4,096) |
| **RoPE Theta** | 500,000 | 1,000,000 | Standard |
| **Pretraining Tokens** | 15 trillion | 18 trillion | Undisclosed |
| **Activation Function** | SwiGLU | SwiGLU | SwiGLU |

**Table S3.** Complete Hyperparameter Configuration

| Parameter | Llama 3.1 8B | Qwen 2.5 7B | Mistral 7B v0.3 |
|---|---|---|---|
| **LoRA Rank (r)** | 16 | 16 | 16 |
| **LoRA Alpha (α)** | 32 | 32 | 32 |
| **LoRA Dropout** | 0.1 | 0.1 | 0.1 |
| **LoRA Target Modules** | Q,K,V,O,Gate,Up,Down | Q,K,V,O,Gate,Up,Down | Q,K,V,O,Gate,Up,Down |
| **Quantization** | 4-bit NF4 | 4-bit NF4 | 4-bit NF4 |
| **Double Quantization** | Enabled | Enabled | Enabled |
| **Compute Dtype** | BFloat16 | BFloat16 | BFloat16 |
| **Learning Rate** | $2\times10^{-4}$ | $2\times10^{-4}$ | $2\times10^{-4}$ |
| **LR Schedule** | Linear + Warmup | Linear + Warmup | Linear + Warmup |
| **Warmup Ratio** | 0.06 | 0.06 | 0.06 |
| **Optimizer** | AdamW | AdamW | AdamW |
| **Weight Decay** | 0.01 | 0.01 | 0.01 |
| **Batch Size (per-device)** | 4 | 4 | 6 |
| **Gradient Accumulation** | 8 | 8 | 5 |
| **Effective Batch Size** | 32 | 32 | 30 |
| **Epochs** | 5 | 5 | 5 |
| **Early Stopping Metric** | Token F1 | Token F1 | Token F1 |
| **Early Stopping Patience** | 3 epochs | 3 epochs | 3 epochs |
| **Max Sequence Length** | 4096 | 4096 | 4096 |
| **Gradient Checkpointing** | Enabled | Enabled | Enabled |
| **Mixed Precision** | Yes (BF16) | Yes (BF16) | Yes (BF16) |

| Random Seed | 42 | 42 | 42 |
|---|---|---|---|
| GPU Memory (Peak) | ~12–14 GB | ~12–14 GB | ~12–14 GB |

**Table S4.** Per-Epoch Performance for Llama 3.1 8B

| Epoch | Train Loss | Val Loss | Generalization Gap | Token F1 | ROUGE-L | BLEU | BERTScore |
|---|---|---|---|---|---|---|---|
| 1 | 1.52 | 1.89 | 0.37 | 0.550 | 0.521 | 0.351 | 0.908 |
| 2 | 1.10 | 1.85 | 0.75 | 0.599 | 0.560 | 0.389 | 0.921 |
| 3 | 1.04 | 1.80 | 0.76 | 0.621 | 0.579 | 0.410 | 0.924 |
| 4 | 0.95 | 1.75 | 0.80 | 0.630 | 0.581 | 0.412 | 0.923 |
| 5 | 0.83 | 1.76 | 0.93 | 0.628 | 0.581 | 0.413 | 0.924 |

**Table S5.** Model ranking after RAG implementation

| Dimension | #1 | #2 | #3 |
|---|---|---|---|
| Accuracy (Token F1) | Mistral 0.737 | Llama 0.686 | Qwen 0.587 |
| Sequence Preservation (ROUGE-L) | Mistral 0.740 | Llama 0.698 | Qwen 0.601 |
| N-gram Overlap (BLEU) | Llama 0.480 | Mistral 0.469 | Qwen 0.346 |
| Semantic Equivalence (BERTScore) | Mistral 0.951 | Llama 0.945 | Qwen 0.933 |
| RAG Improvement | Mistral +195% | Llama +159% | Qwen +143% |
| Speed | Qwen 20.6s | Mistral 25.4s | Llama 28.6s |
| Effect Size | Mistral d=2.8 | Llama d=2.5 | Qwen d=2.1 |
| Consistency | Llama (std=0.031) | Qwen (0.034) | Mistral (0.035) |
| Overall Winner | Mistral (3/4 metrics) | Llama (balanced) | Qwen (fastest) |

**Table S6.** Blind validation scores for the three fine-tuned RAG models evaluated against six mechanistically non-trivial questions derived from unseen, recently published Mg alloy corrosion literature. Each example was scored across three sub-dimensions: environmental/condition classification (A), trend ranking (B), and mechanistic explanation (C), each awarded 0–2 points, yielding a maximum score of 6 per example. Colour coding indicates performance level: green (≥ 4/6), amber (3/6), and red (≤ 2/6). Total scores are out of 36.

| Example | Mechanistic Theme | Llama | Qwen | Mistral |
|---|---|---|---|---|
| 1 | BSA vs SCC/PDP | 5 | 5 | 4 |
| 2 | Time-scale reasoning | 3 | 4 | 1 |
| 3 | AZ-series paradox | 3 | 4 | 1 |
| 4 | EIS state-variable | 3 | 1 | 1 |
| 5 | FSP grain-corrosion | 5 | 2 | 3 |
| 6 | Film thickness paradox | 5 | 4 | 0 |
| Total | | 24/36 | 20/36 | 10/36 |

**Table S7.** Ground truth electrochemical data for Mg-Ca alloy in HBSS at 37°C across four immersion timepoints. |Z| values at 0.01 Hz and OCP end values are reported. Rs = high-frequency solution resistance intercept.

| Timepoint | \|Z\|0.01 Hz (Ω) | Final OCP (V vs ref) | OCP shift during 1 h | Peak phase angle | High-frequency \|Z\| (≈ Rs, Ω) |
|---|---|---|---|---|---|
| 30 min | 3951 | -1.6250 | +87.3 mV | 48.4° @ 12.6 Hz | 62.2 |
| 1 day | 6912 | -1.5703 | -2.7 mV | 50.5° @ 10.0 Hz | 44.3 |
| 5 days | 5706 | -1.6641 | -20.6 mV | 47.3° @ 6.3 Hz | 64.5 |
| 7 days | 3310 | -1.8338 | -0.1 mV | 48.2° @ 12.6 Hz | 45.2 |

**Table S8**. Validation scores (out of 3 per query, 9 total) for LLaMA, Qwen, and Mistral on the three experimental electrochemical queries.

| Query | LLaMA | Qwen | Mistral |
|---|---|---|---|
| **Q1**: **Non-monotonic impedance trajectory** | 3/3 | 1/3 | 2/3 |
| **Q2**: **OCP evolution across immersion time** | 1/3 | 3/3 | 2/3 |
| **Q**3 : **Interpretation of the 7-day electrochemical state** | 3/3 | 1/3 | 2/3 |
| **TOTAL** | **7/9** | **5/9** | **6/9** |

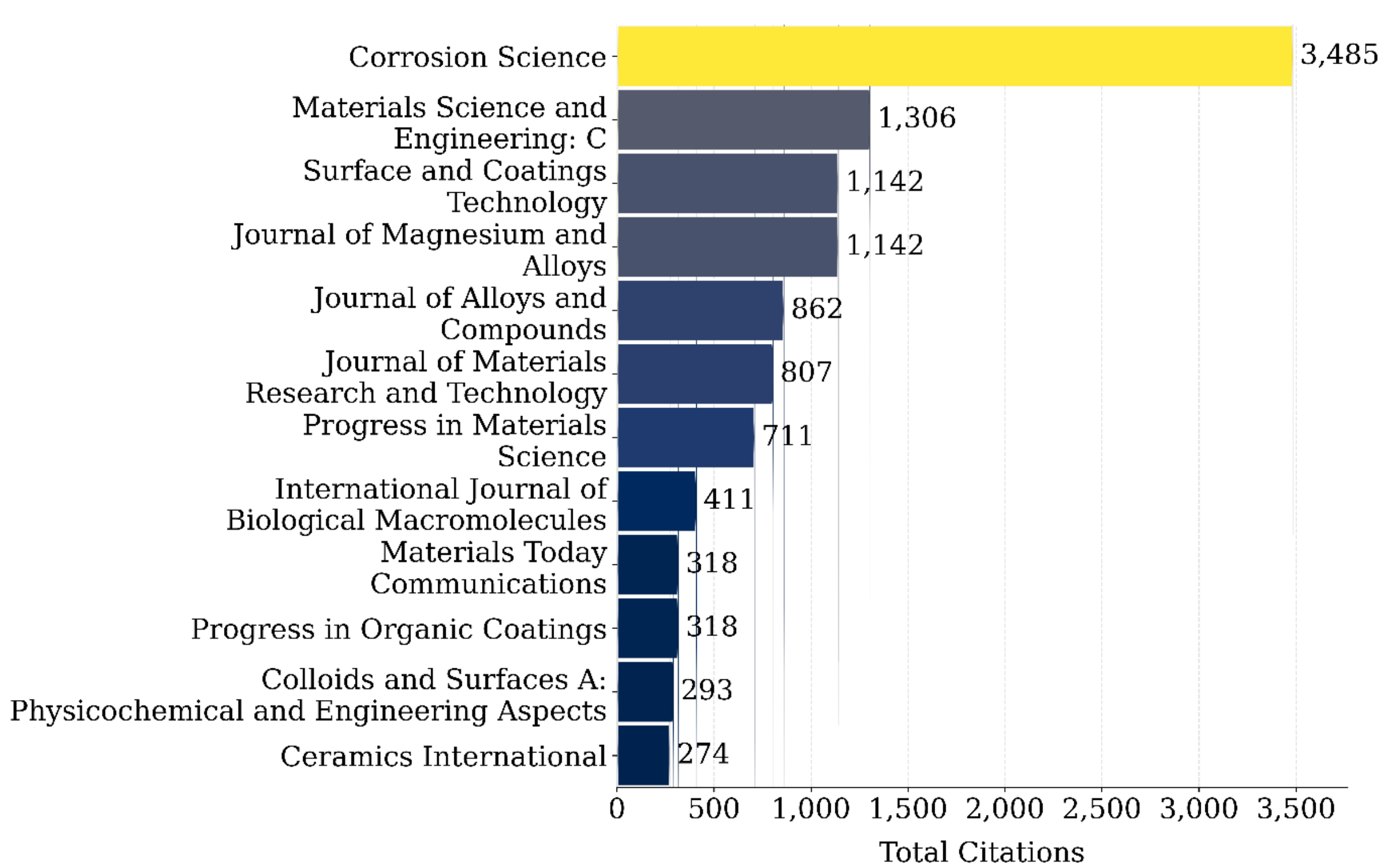

**Figure S1.** Top journals ranked by cumulative citation impact, showing articles from the Corrosion Science journal make the majority of the corpus used for the present study.

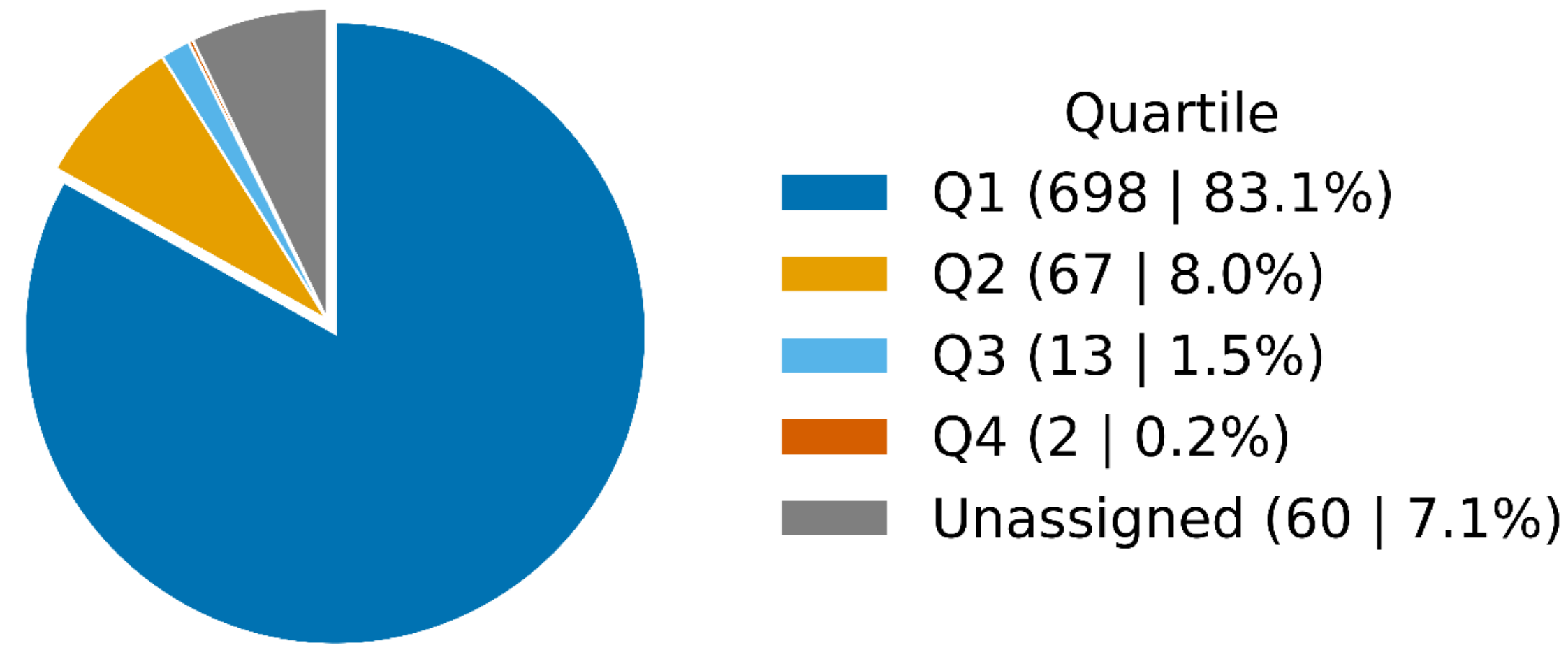

**Figure S2. Distribution of publications across journal quartiles used for the current study.**
The majority of articles fall within Q1 journals (83.1%), with smaller proportions in Q2 (8.0%), Q3 (1.5%), and Q4 (0.2%), while 7.1% remain unassigned.

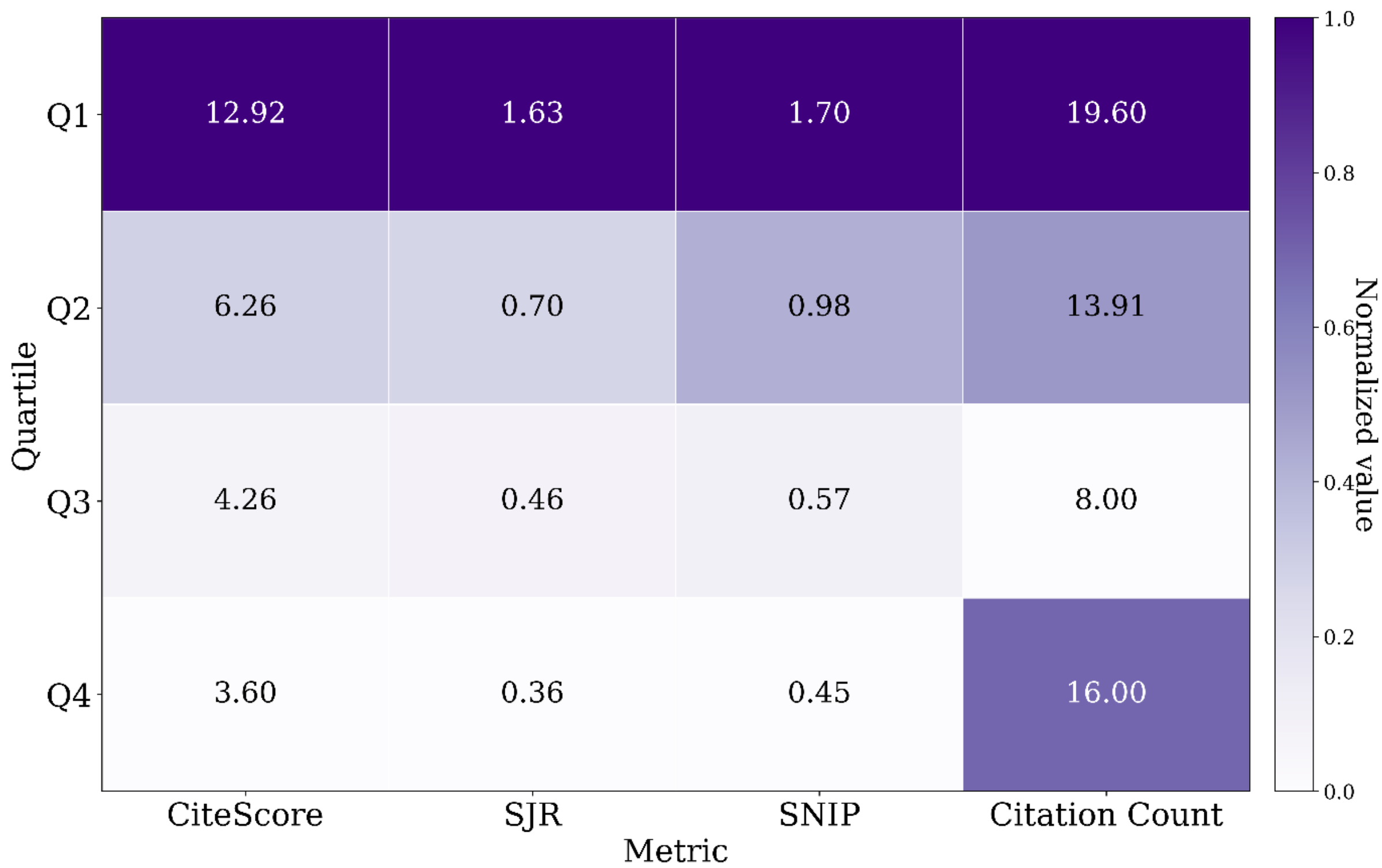

**Figure S3. Heatmap of impact metrics across journal quartiles.**
Mean CiteScore, SJR, SNIP, and Citation Count for Q1–Q4 journals. Colors represent normalized values (0–1) computed separately for each metric to support visual comparison across scales, while the numbers indicate the actual averaged values. Q1 journals outperform lower quartiles across all four indicators.

{
"id": "doc_141_q_011",
"category": "experimental",
"difficulty": "advanced",
"question_type": "procedural",
"query": "What are the critical post-processing treatments required for additively manufactured magnesium scaffolds to achieve optimal biomedical performance?",
"answer": "Critical post-processing treatments for AM magnesium scaffolds include de-powdering, heat treatment, surface modification, and sintering procedures essential for biomedical applications. De-powdering removes unbound particles through compressed air, vacuuming, brushing, and vibration, followed by drying treatment to enhance structural integrity. Heat treatment regimens (T4, T5, T6) enable solid solution hardening, precipitation hardening, homogenization, and stress relief. WE43 alloy processed via FSAM achieves 84.0±3.0 HV baseline hardness, increasing to 94.4±3.0 HV after heat treatment, though T5 condition reaches 103.0±5.0 HV maximum values. Sintering treatment requires temperatures T/T_m > 0.5 for effective powder consolidation, with microwave sintering providing uniform heating to control grain growth. Surface treatments including chemical etching, electrochemical polishing, sandblasting, and plasma electrolytic oxidation (PEO) optimize surface roughness and biological response. Chemical conversion processes using specific solutions create protective surface phases like $MgF_2$ coatings. Hot isostatic pressing (HIP) combines sintering with compaction for enhanced densification. These treatments collectively address thermal stress reduction, microstructural optimization, surface quality enhancement, and biological property improvement necessary for successful clinical implementation of biodegradable magnesium implants.",
"source_section": "Post-processing of Mg scaffolds",
"key_terms": ["post-processing", "heat treatment", "surface modification", "sintering", "plasma electrolytic oxidation"],
"quantitative_data": true,
"environmental_conditions": ["controlled temperature treatment", "specific atmosphere conditions"],
"material_specificity": ["WE43", "FSAM processed", "various AM techniques"],
"biomedical_context": "orthopedic",
"domain": "biomedical",
"research_focus": "experimental",
"relevance_score": 4,
"validation_notes": "Comprehensive post-processing procedures with quantitative hardness data and specific treatment conditions"
},

**Figure S4. Example of a generated and validated Q&A pair format.**
Schema-rich Q&A record demonstrating the structure of training/evaluation items: unique ID, category/difficulty/question type, natural-language query, and vetted answer containing quantitative details; followed by structured metadata. Records are stored line-by-line in JSONL to enable scalable streaming during fine-tuning and reproducible audits.

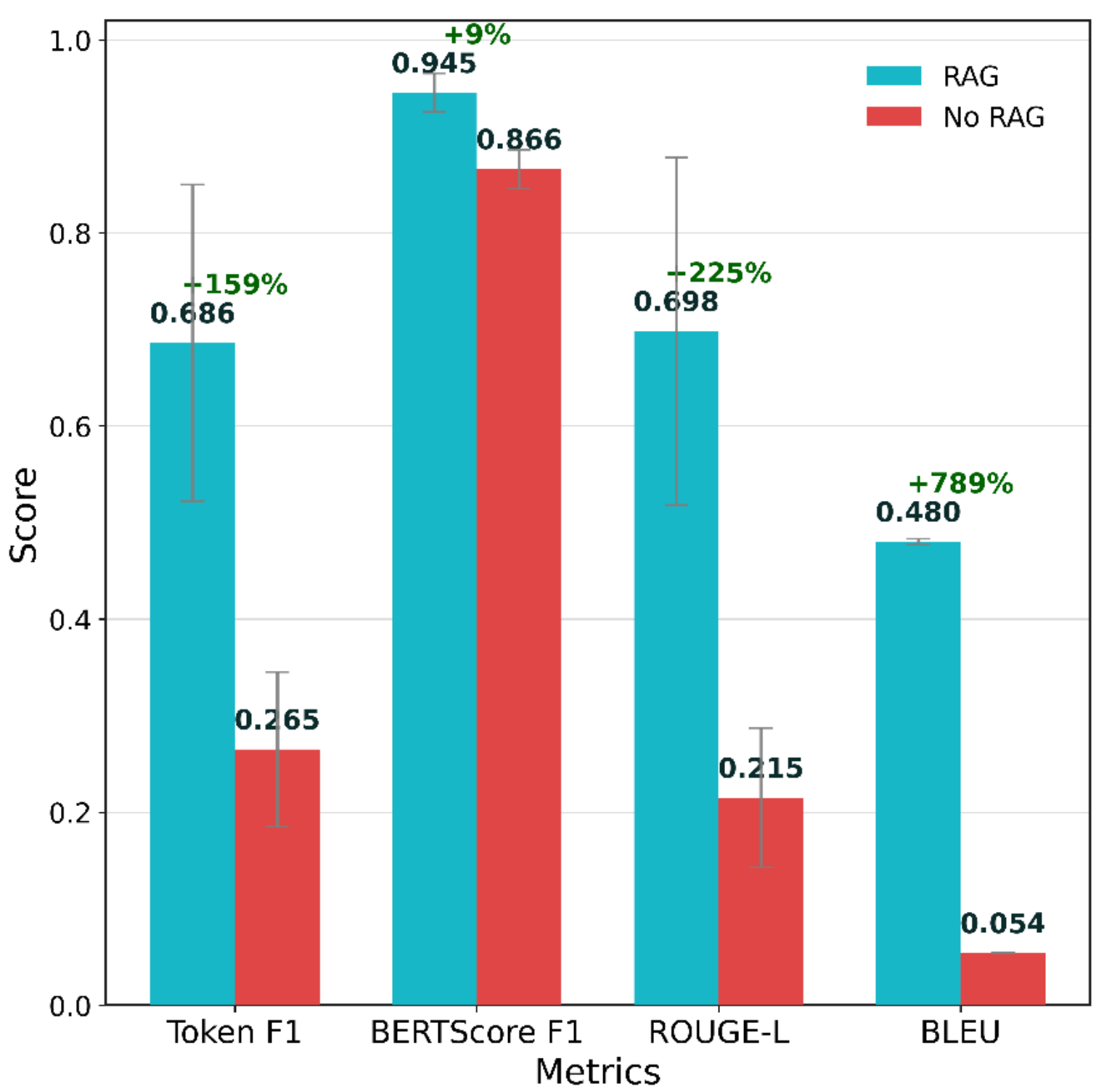

**Figure S5. Performance of Llama-3.1-8B instruct with and without RAG across four metrics.**
Mean scores (bars) and one standard deviation (error bars) for token F1, BERTScore F1, ROUGE-L and BLEU under the RAG (blue) and no-RAG (red) conditions on 140 Mg-alloy corrosion questions. RAG consistently outperforms the baseline across all metrics, indicating more faithful and better structured answers when retrieved evidence is available.

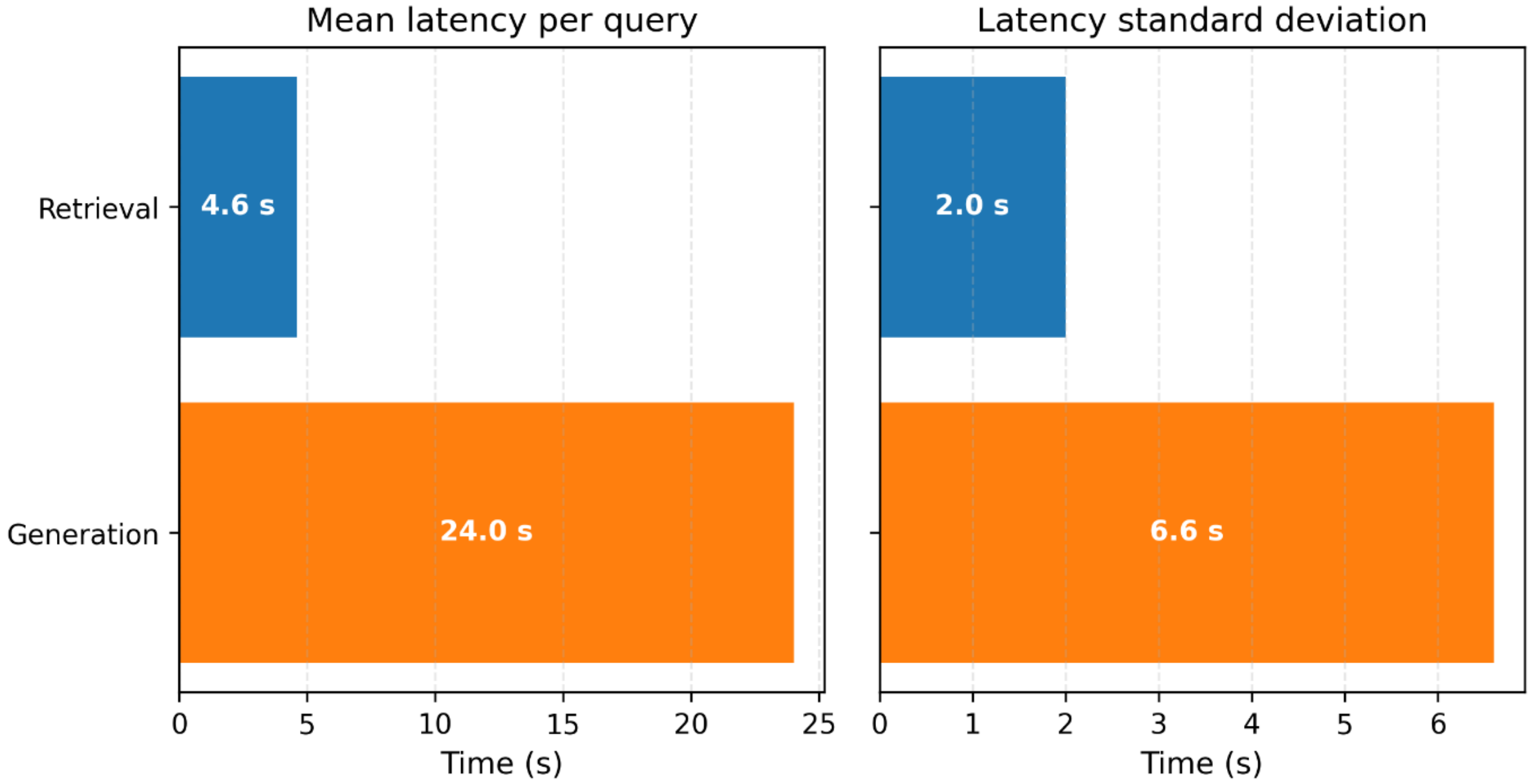

**Figure S6. Latency profile of the Llama-3.1-8B RAG system.**
Mean latency per query (left) and standard deviation of latency across queries (right) for the retrieval and generation stages on our evaluation hardware. Retrieval takes on average 4.6 s per query, whereas LLM generation takes 24.0 s, for a total of ≈28.6 s. The corresponding standard deviations are 2.0 s and 6.6 s, indicating that generation is both slower and more variable than retrieval.

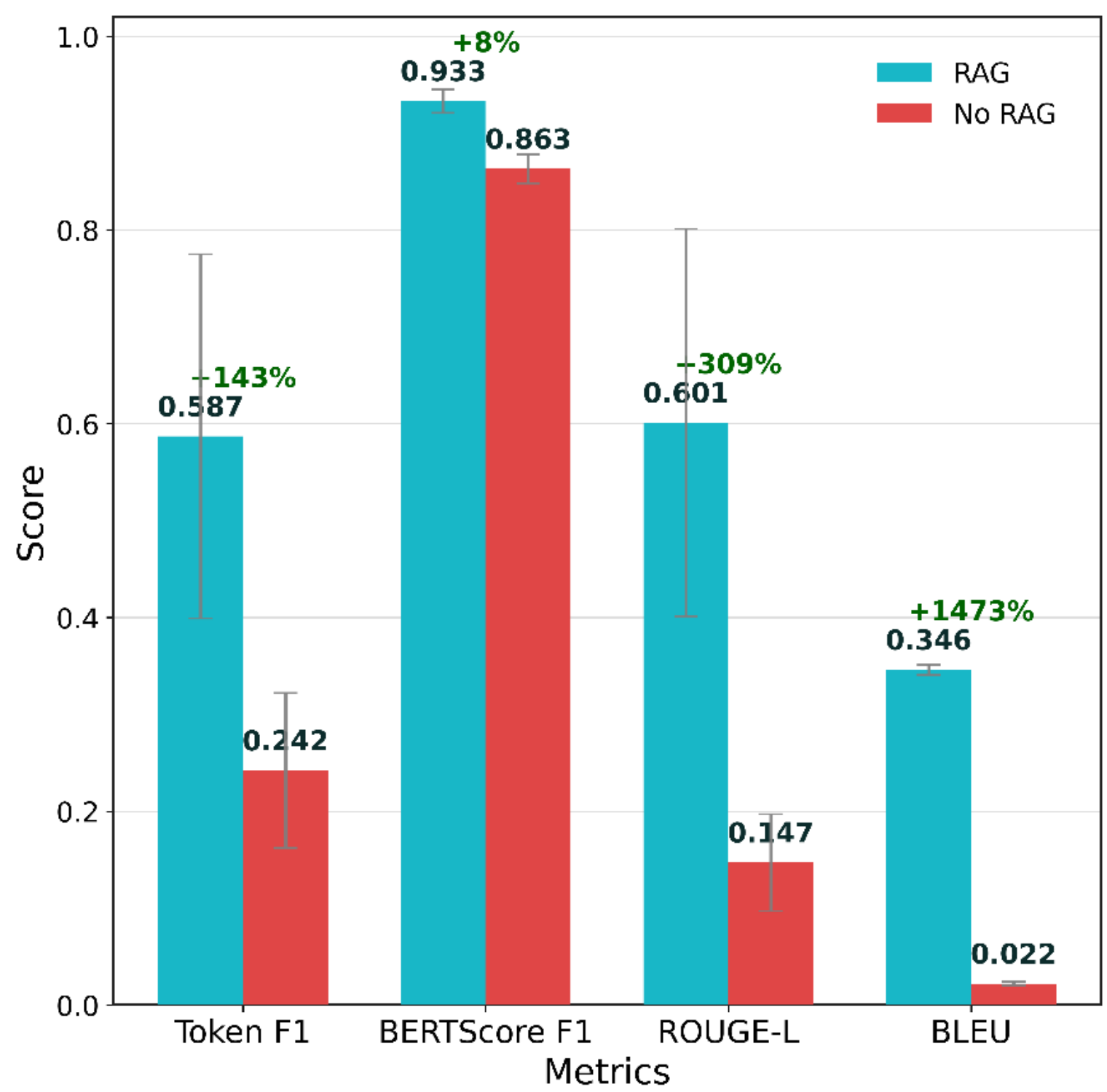

**Figure S7. Performance of the fine-tuned Qwen-2.5-7B model with and without RAG.**
Mean scores over 140 Mg-alloy corrosion questions for token F1, BERTScore F1, ROUGE-L and BLEU. Teal bars show the RAG configuration (model conditioned on retrieved papers and FAQs); red bars show the no-RAG baseline (question only). Error bars denote one standard deviation across test questions. Numeric labels give the mean score for each condition, and green annotations indicate the relative improvement (%) of RAG over the baseline.

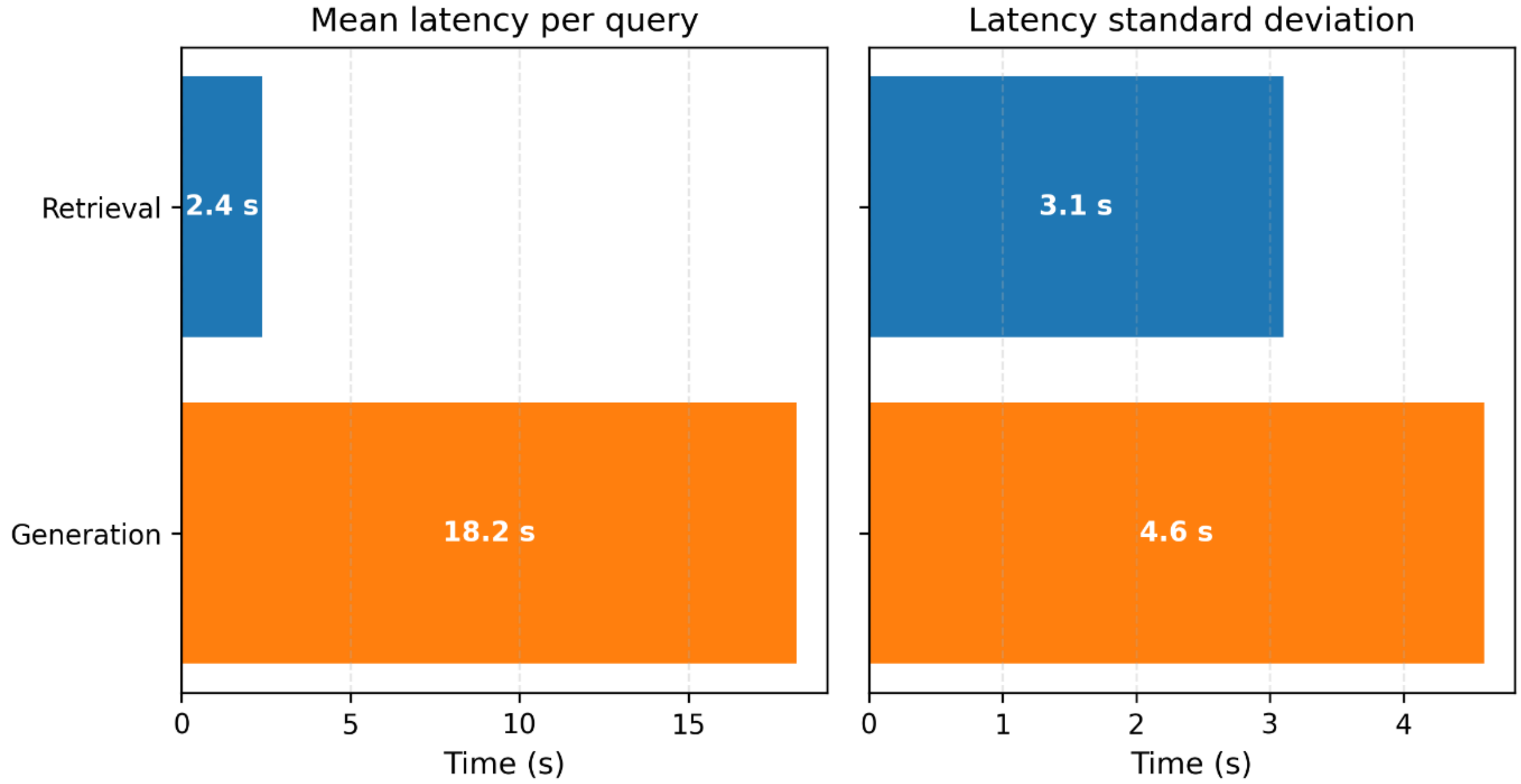

**Figure S8. Latency profile of the Qwen-2.5-7B RAG system.**
Mean latency per query (left) and standard deviation across queries (right) for the retrieval and generation stages when answering Mg-alloy corrosion questions with Qwen-2.5-7B. Hybrid retrieval takes on average 2.4 s per query with a standard deviation of 3.1 s, reflecting substantial variability in the cost of reranking. Generation dominates the total response time, requiring 18.2 ± 4.6 s per query. The combined mean latency is therefore 20.6 s per query in our unoptimised research implementation.

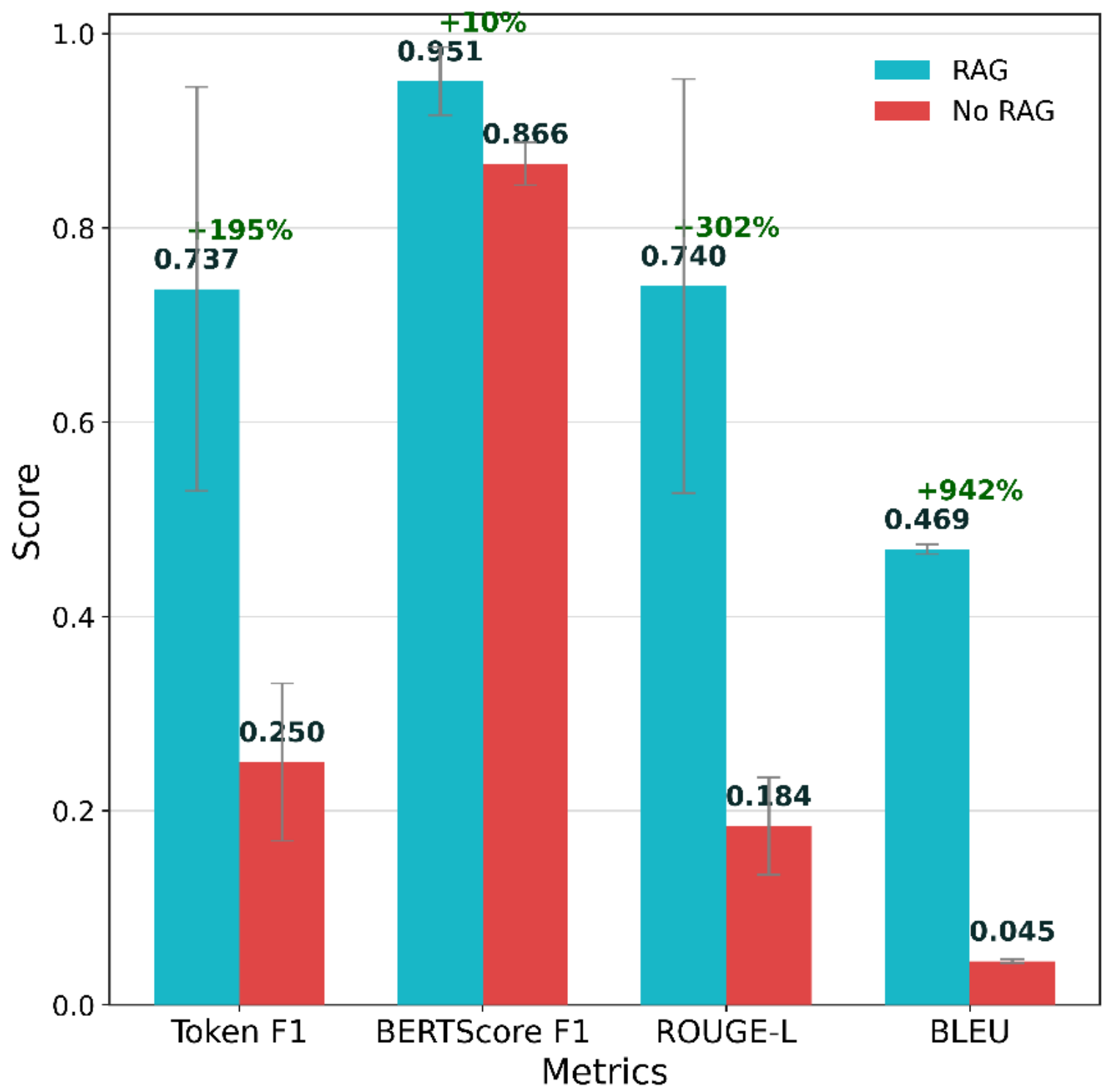

**Figure S9. Performance of the fine-tuned Mistral-7B model with and without RAG.**
Mean scores over 140 Mg-alloy corrosion questions for Token F1, BERTScore F1, ROUGE-L and BLEU. Teal bars show the RAG configuration (model conditioned on retrieved papers and FAQs); red bars show the no-RAG baseline (question only). Error bars denote one standard deviation across test questions. With RAG, Mistral achieves Token F1 = 0.737, BERTScore F1 = 0.951, ROUGE-L = 0.740 and BLEU = 0.469, compared with 0.250, 0.866, 0.184 and 0.045 for the baseline, indicating very large absolute and relative gains across all metrics.

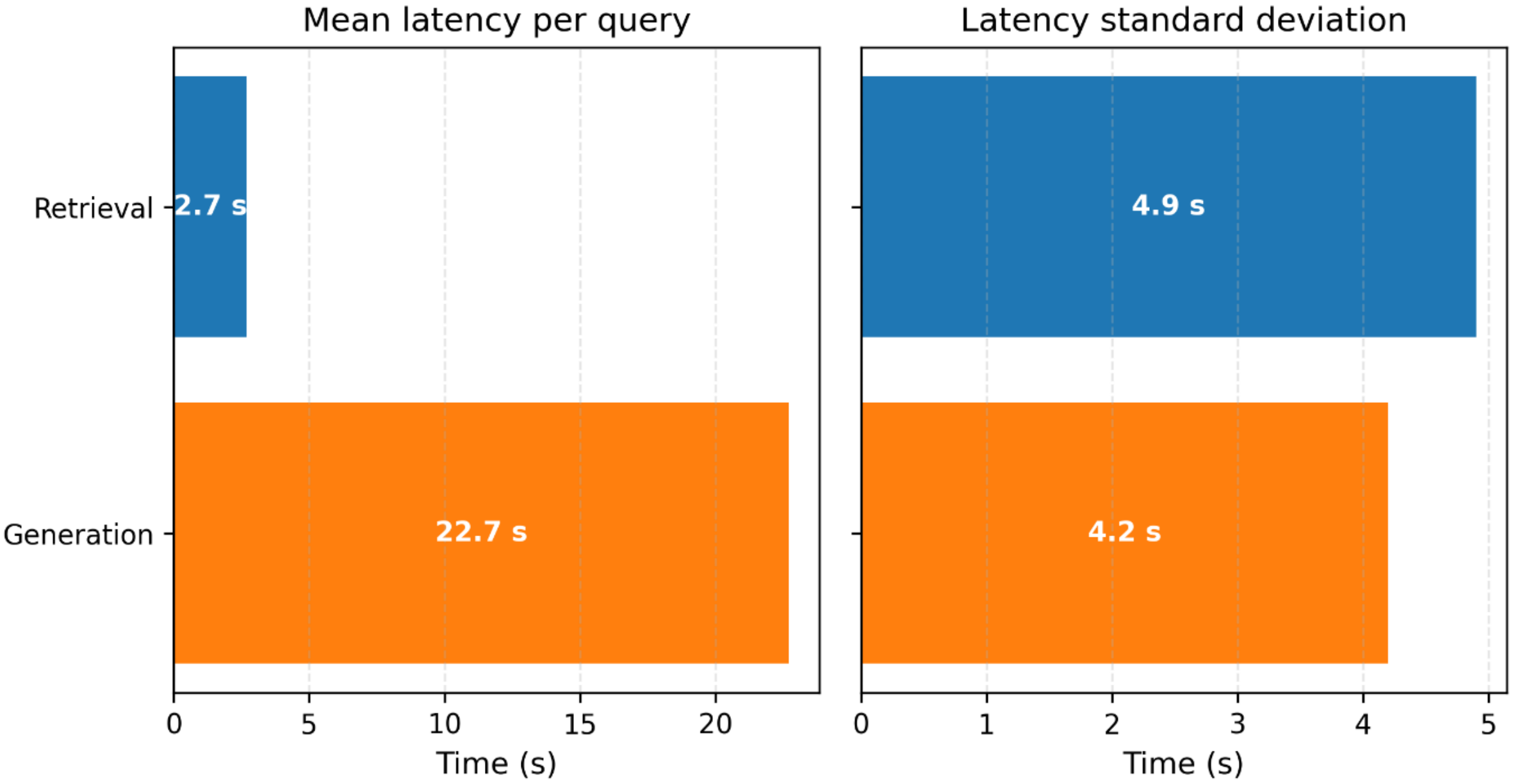

**Figure S10. Latency profile of the Mistral-7B RAG system.**
Mean latency per query (left) and standard deviation across queries (right) for the retrieval and generation stages when answering Mg-alloy corrosion questions with Mistral-7B. Hybrid retrieval requires 2.7 ± 4.9 s per query, while generation takes 22.7 ± 4.2 s, giving a total mean response time of 25.4 s. Retrieval thus accounts for only ~11% of the wall-clock time, with generation dominating the remaining ~89%.

Materials Science RAG Assistant

Advanced Question Answering for Magnesium Alloys and Corrosion

Powered by hybrid retrieval (BM25 + Dense + RRF + Reranking) and fine-tuned LLMs.

Ask Question | Compare Models | Reason Map | About

Your Question

How does protein influences Mg alloy corrosion?

Select Model

Llama 3.1 8B

Submit

Example Questions

How does laser surface remelting (LSR) affect AZ31B corrosio...

What are the main corrosion mechanisms in magnesium alloys?

Compare the corrosion performance of AZ31 and AZ91 alloys.

What is the role of rare earth elements in Mg alloy corrosio...

How does microstructure influence corrosion behavior in Mg-R...

Settings

Top-K Contexts 5 (1–10)

Include FAQs

Rerank with Cross-Encoder

Use HyDE

Max Answer Length 512 (128–1024)

Show Full Context

Use sampling (temperature/top-p)

Temperature 0.7 (0.1–1.5)

Top-p 0.9 (0.1–1)

Protein influences Mg alloy corrosion through a synergistic effect of adsorption, cation bonding, and interfacial pH control. Protein adsorption on the Mg surface creates a protective layer that can inhibit corrosion, while cation bonding between protein and Mg2+ ions promotes dissolution of the Mg matrix, accelerating corrosion. The buffering effect of protein also accelerates corrosion by maintaining a stable pH environment. The combined effect results in uniform corrosion with fewer pits, but mechanical properties decrease slowly and steadily.

Performance:

- Retrieval: 0.66s
- Generation: 3.60s
- Total: 4.26s
- Model: Llama 3.1 8B
- Retrieved: 5 contexts
- do_sample: False | temperature: 0.70 | top_p: 0.90

*Switch to the **Reason Map** tab to validate this answer's reasoning.*

Retrieved Contexts

Full Context (for debugging)

**Note**: A GPU is recommended for best performance. CPU inference is supported but slower.

**Figure S11. Materials Science RAG Assistant web interface- Ask Question tab.**
The representative query *"How does protein influence Mg alloy corrosion?"* submitted to Llama 3.1 8B illustrates the configurable retrieval settings panel (Top-*K*, cross-encoder reranking, HyDE, answer length, sampling controls), the generated answer with inline performance statistics (retrieval: 0.66 s, generation: 3.60 s, total: 4.26f s), and expandable retrieved context display with source attribution. The interface prompt at the bottom of the statistics panel directs users to the Reason Map tab for proposition-level reasoning validation of the generated answer

**Supplementary Note S1: Model Validation using an external paper, excluded from both training and indexing.**

The three RAG models were evaluated on questions derived from three recently published magnesium alloy corrosion papers, entirely excluded from both training and indexing. Six mechanistically non-trivial examples were evaluated blind.

**Paper selected for validation:**

RK Singh Raman, Amal Sibi, Dandapani Vijayshankar, M.J.N.V. Prasad, G. Keerthiga, Solomon Ansah, Saad Al-Saadi, Jafar Albinmousa, Protein in physiological fluid resists premature fracture of a magnesium alloy: Unique, remarkable and contrasting influences on stress corrosion cracking and corrosion, Journal of Magnesium and Alloys, Volume 13, Issue 12, 2025, Pages 5842-5854, ISSN 2213-9567, https://doi.org/10.1016/j.jma.2025.08.036.

**Example 1- SCC vs PDP contradiction (external paper validation)**

**Question asked to models (blind; no paper uploaded):**

"Consider a Mg alloy (ZK60-type) tested in Hanks' solution at 37 °C under SSRT for SCC, and also tested by potentiodynamic polarization after long OCP stabilization. Four environments: (HS1) Hanks, (HS2) Hanks + ~1 g/L glucose, (HS3) Hanks + ~40 g/L albumin/protein (BSA), (HS4) HS2 + BSA. Answer with trends only and a short mechanism.

Same setup. Answer in 3 bullet points only.

1. During SSRT SCC, which condition(s) should show SCC features (secondary cracks / TG-IG cracking) and which should show mechanical overload?
2. During PDP after long OCP stabilization, in which condition(s) should icorr be higher?
3. Give one mechanism that explains "SCC decreases but icorr increases" with BSA, explicitly referencing fresh crack-tip film vs long-exposure film evolution."

**LLAMA response:**

"1. SCC features (secondary cracks/TG-IG cracking) are expected in HS1 and HS2, while mechanical overload (ductile dimples) is expected in HS3 and HS4 due to protein adsorption and film formation.

2. icorr is higher in HS3 and HS4 compared to HS1 and HS2, indicating increased corrosion rate.

3. The mechanism is that fresh crack-tip films in HS3 and HS4 are more protective than long-exposure films in HS1 and HS2, reducing SCC susceptibility but increasing icorr due to increased film resistance."

**Score (5/6):** A=2/2 (SSRT mapping correct), B=2/2 (icorr trend correct), C=1/2 (mentions fresh crack-tip film but misses the paper's key "long-exposure chelation/complex formation disrupts film" explanation).

**MISTRAL response:**

"During SSRT SCC, HS1 (Hanks) shows SCC features, HS2 (Hanks + glucose) shows mechanical overload, HS3 (Hanks + BSA) shows no significant change, and HS4 (Hanks +

glucose + BSA) shows SCC decreases but mechanical overload. During PDP after long OCP stabilization, icorr is higher in HS3 (Hanks + BSA) and HS4 (Hanks + glucose + BSA) due to chelation/complex formation disrupting corrosion film, explaining why SCC decreases despite increased corrosion activity."

**Score (4/6):** A=0/2 (HS2 and HS3 misclassified; paper says HS2 behaves like HS1 and HS3 behaves like HS4), B=2/2 (icorr trend + glucose insignificance captured), C=2/2 (mechanism matches chelation/film disruption).

**QWEN response:**

"SCC features: HS1 (pure Hanks), HS2 (glucose present), HS4 (glucose + BSA). Mechanical overload: HS3 (pure BSA). Mechanism explanation: Initially, BSA adsorption on crack tips prevents rapid corrosion and delays SCC initiation, reducing secondary crack formation. However, during prolonged immersion, BSA molecules form complexes with corrosion products such as $Mg(OH)_2$ and MgO, disrupting the protective film structure and increasing local ion diffusion rates. This complex formation causes localized pitting and increased corrosion current density (icorr), even as overall SCC damage decreases. The dynamic balance between protective BSA adsorption and film disruption through complex formation explains the observed increase in icorr in BSA-containing solutions after long-term exposure."

**Score (5/6):** A=1/2 (HS4 misclassified; paper says HS4 is similar to HS3 with dimples/overload), B=2/2 (icorr increase with BSA captured), C=2/2 (fresh-film vs chelation mechanism captured well).

---

**Example 2- Time-scale reasoning (minutes vs days)**

**Question asked to models (blind; no paper uploaded):**

"Consider the same HS1–HS4 setup. How might the effect of albumin on Mg corrosion and SCC change from minutes to days of exposure? Explain."

**QWEN response:**

"Initially, albumin (HS3) would decrease corrosion current density and delay SCC crack initiation by stabilizing protective layers, showing beneficial effects within minutes. Over days, albumin's protective role diminishes due to protein denaturation, aggregation, and loss of structural integrity, allowing chloride ingress and pit formation. This transition from short-term inhibition to long-term acceleration explains why some studies report initial protection with albumin but increased corrosion later. Mechanistically, albumin adsorption forms compact films initially, but over time proteins undergo conformational changes and cross-linking, reducing barrier effectiveness."

**Score (4/6):** A=2/2 (correct "early protection → late acceleration" direction), B=1/2 (early SCC delay stated, but doesn't clearly connect to "fresh crack-tip film" SCC mechanism), C=1/2 (paper emphasizes **chelation/complex formation and film removal**, not denaturation/crosslinking).

**LLAMA response:**

"Initially, albumin (BSA) accelerates Mg corrosion by forming a protective protein layer that hinders further degradation. However, over time, the protective layer breaks down, and albumin's chelating properties become dominant, further accelerating corrosion. This dual effect explains why albumin's influence on corrosion and SCC changes from protective to detrimental with prolonged exposure."

**Score (3/6):** A=1/2 (contradictory phrasing: 'accelerates corrosion by forming a protective layer'), B=1/2 (mentions SCC qualitatively but not the paper's "fresh crack-tip film controls SCC"), C=1/2 (does mention chelation, which is central to the paper's long-exposure corrosion explanation).

**MISTRAL response:**

"Initially, albumin acts as a corrosion inhibitor by adsorbing to the Mg surface, reducing corrosion rates and stabilizing the corrosion product layer. However, over time (days), albumin promotes more uniform corrosion and delays the decline of mechanical properties, but does not significantly reduce corrosion rates compared to Hanks solution alone. This is because albumin adsorption provides temporary protection that wears off as corrosion progresses, allowing uniform corrosion to dominate. In SCC testing, albumin initially stabilizes corrosion pits, but over time these pits become more uniform and act as crack initiation sites, potentially accelerating crack propagation. Thus, albumin's effect on corrosion and SCC changes from inhibition to promotion over extended exposure times."

**Score (1/6):** A=1/2 (time dependence noted but misses the paper's "BSA increases corrosion susceptibility after long exposure"), B=0/2 (predicts SCC worsening; paper concludes absence of SCC in BSA conditions), C=0/2 (missing chelation/complex formation and fresh crack-tip film framing).

---

**Paper selected for validation:**

Jingying Li, Hao Liu, Li Liu, Chongqing Guo, Lihui Yang, Jiarun Li, Jia Wang, Effect of aluminum and zinc on the corrosion behavior of AZ (Mg-Al-Zn) series magnesium alloys, Journal of Materials Science & Technology, Volume 246, 2026, Pages 116-130, ISSN 1005-0302, https://doi.org/10.1016/j.jmst.2025.04.007.

**Example-3: AZ63 vs AZ91 corrosion paradox**

**Question asked to models (blind; no paper uploaded):**

Consider five as-cast AZ magnesium alloys exposed to 3.5 wt.% NaCl at 25 °C: AZ31, AZ61, AZ62, AZ63, and AZ91.

The alloys differ mainly in Al/Zn content and in the relative amounts/morphologies of β-$Mg_{17}Al_{12}$ and eutectic α phases.

Answer in 3 bullet points only. Use trends only plus one short mechanism.

1. Rank the alloys for overall corrosion resistance after immersion/electrochemical testing from best to worst.
2. Which alloy should show the most localized corrosion, and which alloy should show the most uniform but fastest corrosion?
3. Give one mechanism that explains why an alloy with less β-$Mg_{17}Al_{12}$ can still corrode faster than one with more β-$Mg_{17}Al_{12}$, explicitly referencing eutectic α, micro-galvanic effects, and corrosion-product desorption/fresh surface exposure.

**Gold answer from the paper:**
Overall corrosion resistance, from best to worst, is AZ31 > AZ61 > AZ62 > AZ91 > AZ63. The immersion data show AZ31 lowest and AZ63 highest corrosion rate, with AZ61 better than AZ62 and AZ91 worse than both; the electrochemical data are consistent, with Rp decreasing from AZ31 to AZ63 and AZ91 lower than AZ62. The paper also states that AZ91 shows the most serious localized/micro-galvanic corrosion, whereas AZ63 shows the most uniform corrosion while still having the highest overall corrosion rate. The key explanation is that AZ63 has more eutectic α, which buffers severe β/α galvanic attack and makes corrosion more uniform, but the corrosion products and even discrete β phases desorb more readily, exposing fresh Mg surface; by contrast, AZ91 has more β-$Mg_{17}Al_{12}$, but its network can exert an anodic barrier effect and tends to retain corrosion products, so it is highly localized but not the fastest overall.

**Llama:** • AZ31 shows the best overall corrosion resistance due to balanced Al and Zn content, with minimal β-$Mg_{17}Al_{12}$ and eutectic α phases. • AZ91 exhibits the worst corrosion resistance due to high Zn content causing excessive eutectic α phase formation and severe micro-galvanic effects between α-Mg and β-$Mg_{17}Al_{12}$ phases. • AZ63 shows the fastest corrosion rate due to high Al content causing excessive β-$Mg_{17}Al_{12}$ formation, which promotes micro-galvanic corrosion and facilitates desorption of corrosion products, exposing fresh surfaces for further attack. • Mechanism: Although AZ63 has more β-$Mg_{17}Al_{12}$, its corrosion rate is higher than AZ91 due to the presence of eutectic α phases that corrode preferentially, creating fresh surfaces for further attack. The β-$Mg_{17}Al_{12}$ in AZ63 acts as a cathode, accelerating corrosion of the α-Mg matrix, while in AZ91, the eutectic α phases dominate, leading to more uniform but slower corrosion.
**Qwen:** Corrosion resistance ranking: AZ91 < AZ63 < AZ62 = AZ61 < AZ31. Localized corrosion is most severe for AZ91 due to abundant eutectic β-$Mg_{17}Al_{12}$. Uniform but fastest corrosion occurs for AZ63 despite β-$Mg_{17}Al_{12}$ presence. Mechanism: In AZ63, increased eutectic β-$Mg_{17}Al_{12}$ promotes cathodic hydrogen evolution, accelerating local anodic dissolution of α-Mg near β-phase networks. Simultaneously, corrosion product desorption exposes fresh α-Mg surfaces, creating continuous corrosion propagation. This synergy between cathodic activity and fresh surface exposure leads to higher overall corrosion rates despite some β-phase barrier effects.
**Mistral:** Ranking: AZ31 > AZ62 > AZ61 > AZ63 > AZ91. AZ91 should show the most localized corrosion due to higher β-$Mg_{17}Al_{12}$ content and eutectic α phase distribution. AZ31 corrodes fastest despite less β-$Mg_{17}Al_{12}$ because eutectic α phase promotes uniform corrosion, desorption of corrosion products exposes fresh surfaces, and the cathodic process is strongly promoted by Al and Zn.

**LLAMA — Score: 3/6**

A = 1/2: It gets AZ31 as best and AZ63 as very poor/fast, but it wrongly makes AZ91 the worst and confuses the chemistry by attributing AZ91 mainly to “high Zn,” whereas AZ91 is the high-Al alloy and AZ63 is the Zn-rich one.
B = 1/2: It hints that AZ63 is the fastest, but it does not cleanly identify AZ91 as the most localized and AZ63 as the most uniform-fast in the way the paper shows.
C = 1/2: It at least mentions eutectic α preferential corrosion and fresh-surface exposure, but it incorrectly says AZ63 has more β-$Mg_{17}Al_{12}$ than AZ91, which is opposite to the paper’s core AZ63-vs-AZ91 paradox.

**QWEN — Score: 4/6**
A = 1/2: Better than Llama on the broad trend, but the ranking is still wrong because the paper supports AZ63 as worst, not AZ91, and AZ61 better than AZ62, not equal.
B = 2/2: It correctly identifies AZ91 as most localized and AZ63 as uniform but fastest, which is one of the main non-trivial conclusions of the paper.
C = 1/2: It captures fresh surface exposure and mentions some β-phase barrier effect, but it misses the paper’s specific mechanistic emphasis that eutectic α, not “eutectic β,” is the crucial Zn-driven feature behind the AZ63 behavior.

**MISTRAL — Score: 1/6**
A = 0/2: The ranking is substantially wrong.
B = 1/2: It correctly says AZ91 is most localized, but it completely misses the “uniform but fastest = AZ63” part and instead says AZ31 corrodes fastest, which is directly contradicted by the paper.
C = 0/2: The mechanism is assigned to the wrong alloy, so even though terms like eutectic α and desorption appear, the reasoning does not match the paper’s actual AZ63-vs-AZ91 explanation.

**Example 4: AZ91 EIS state-variable reasoning under OCP vs ±120 mV polarization**

**Question asked to models (blind; no paper uploaded):**

Consider an AZ91 magnesium alloy immersed in 3.5 wt.% NaCl for 24 h. EIS is measured in three conditions: at free corrosion potential, at **+120 mV anodic polarization**, and at **−120 mV cathodic polarization**.

Answer in **3 bullet points only**. Use **features/trends only** and one short mechanism.

1. At free corrosion potential, which combination of loops should appear in the Nyquist plot?
2. When moving to **strong anodic** versus **strong cathodic** polarization, which loop(s) should **persist** and which should **disappear**?
3. Give one mechanism that explains this behavior, explicitly referencing the two surface state variables: **product-free area fraction (θ)** and **univalent Mg ion activity ($C_{Mg+}$)**, and why one of them no longer contributes under strong cathodic polarization.

**Gold answer from the paper:**

For AZ91 at free corrosion potential, the Nyquist plot shows two capacitive loops and one inductive loop. Under +120 mV anodic polarization, that same pattern remains: two capacitive loops plus one inductive loop. Under −120 mV cathodic polarization, the response collapses to one capacitive loop only. The paper’s mechanism is that the EIS response at free corrosion combines anodic and cathodic processes, but the two extra state variables are anodic-state variables: θ (product-free area fraction), which gives a capacitive response, and CMg+ (univalent Mg ion activity), which gives an inductive response. Under strong cathodic polarization, the anodic dissolution pathway to Mg+ is suppressed and θ no longer varies with potential, so those extra responses disappear, leaving only the double-layer capacitive loop.

**Mistral:** At free corrosion potential, the Nyquist plot should show two capacitive loops and one inductive loop. Under strong anodic polarization (+120 mV), the two capacitive loops persist while the inductive loop disappears due to increased product-free area fraction (θ) and higher univalent Mg ion activity (CMg+) at the surface. Under strong cathodic polarization (-120 mV), only one capacitive loop remains while the inductive loop disappears because the cathodic reaction ($2H_2O + 2e^- \rightarrow H_2 + 2OH^-$) dominates, consuming hydroxide ions and reducing the surface pH, which suppresses the formation of the inductive loop associated with the Mg(OH)2 film.

**Llama:** • At free corrosion potential, two capacitive loops and one inductive loop should appear in the Nyquist plot, corresponding to the inner compact corrosion product film, outer loose corrosion product film, and double layer capacitance/charge transfer resistance. • Under strong anodic polarization (+120 mV), all three loops persist, while under strong cathodic polarization (−120 mV), only one capacitive loop remains, with the inductive loop and double layer capacitance/charge transfer resistance loop disappearing. • The mechanism is that under strong cathodic polarization, the surface is fully covered by corrosion products (θ = 1), and univalent Mg ion activity (CMg+) is suppressed, eliminating the contribution of double layer capacitance and charge transfer resistance to the EIS response.

**Qwen:** At free corrosion potential, the Nyquist plot should display three capacitive loops: one at high frequency (inner compact corrosion product), one at intermediate frequency (outer loose corrosion product), and one at low frequency (double layer and charge transfer). When moving to strong anodic polarization (+120 mV), all three loops persist, indicating active anodic dissolution and product formation. Under strong cathodic polarization (−120 mV), only one capacitive loop remains at high frequency while intermediate and low frequency loops disappear. This behavior is explained by the fact that at strong cathodic polarization, the entire surface is covered with corrosion products (θ=1), eliminating the contribution of product-free area (inner loose corrosion product loop disappears). The dominant mechanism becomes charge transfer and double layer capacitance, with only the outer compact product contributing to capacitive response.

**Mistral — Score: 1/6**

A = 1/2: It gets the free-corrosion case right and correctly says that under strong cathodic polarization only one capacitive loop remains, but it wrongly removes the inductive loop under anodic polarization, whereas the paper shows it still persists at +120 mV.

B = 0/2: It mentions θ and CMg+, but does not assign them correctly as θ → capacitive and CMg+ → inductive.

C = 0/2: The cathodic explanation is off. The paper does not explain the disappearance of the extra loops by hydroxide consumption or pH reduction; it explains it by the loss of the anodic-state-variable responses under strong cathodic polarization.

**Llama — Score: 3/6**

A = 2/2: It correctly captures the loop pattern across the three conditions: two capacitive + one inductive at free corrosion, the same under +120 mV anodic polarization, and one capacitive loop under −120 mV cathodic polarization.

B = 0/2: The loop attribution is wrong. For AZ91, the paper does not reduce the response to “inner film, outer film, and double layer” in the way Llama states, and it misses the intended state-variable mapping $\theta$ → capacitive and CMg+ → inductive.

C = 1/2: It at least recognizes that CMg+ is suppressed under strong cathodic polarization, which is directionally consistent. But it then wrongly says the surface is fully covered by corrosion products and that the double-layer/charge-transfer response disappears, whereas the paper indicates that the remaining loop is precisely the double-layer capacitive response.

**Qwen — Score: 1/6**

A = 1/2: It correctly says that under strong cathodic polarization only one capacitive loop remains, but it misreads the free-corrosion and anodic cases as three capacitive loops, whereas the paper clearly shows two capacitive loops and one inductive loop.

B = 0/2: It does not correctly assign the state variables; the paper’s key point is $\theta$ gives the capacitive response and CMg+ gives the inductive response.

C = 0/2: Its cathodic explanation relies mainly on $\theta$ and product coverage, but it misses the central point that strong cathodic polarization suppresses the anodic Mg+ pathway, so the anodic-state-variable responses vanish.

---

Hou-Jen Chen, Prakash Chandra Gautam, Pi-Chen Lin, Chih-Kai Wang, Hsin-Chih Lin,

Friction stir processing–induced microstructure refinement enhances corrosion resistance of ZE52 magnesium alloy in simulated physiological environment, Journal of Alloys and Compounds, Volume 1057, 2026, 186706, ISSN 0925-8388, https://doi.org/10.1016/j.jallcom.2026.186706.

**Example 5: Finest grains do not equal best corrosion resistance**

**Question asked to models (blind; no paper uploaded):**

Consider a ZE52 magnesium alloy tested in simulated body fluid (SBF) at 37 °C. One condition is the base material (BM), and three conditions are friction stir processed at constant 1200 rpm with traverse speeds of 50, 100, and 200 mm/min.

Answer in 3 bullet points only. Use trends only plus one short mechanism.

1. Which condition should show the finest grain size, and which condition should show the best corrosion resistance by EIS/PDP?
2. Rank BM, 1200–50, 1200–100, and 1200–200 from worst to best corrosion resistance.
3. Give one mechanism that explains why the condition with the very finest grains is not necessarily the one with the best corrosion resistance, explicitly referencing precipitate fragmentation/homogenization, solute re-solution into α-Mg, basal texture, and stable alloy-oxide formation.

**Gold answer from the paper:**

The finest grains are in FSP 1200–100 (about 2.47 ± 0.73 μm), while the best corrosion resistance is in FSP 1200–200, which shows the largest Nyquist loop, the highest Rp (914.1 $\Omega \cdot cm^2$), and the lowest Icorr ($1.26 \times 10^{-5}$ $A \cdot cm^{-2}$). The corrosion-resistance ranking from worst to best is BM → 1200–50 → 1200–100 → 1200–200. The key point is that corrosion resistance is not governed by grain size alone: the best condition combines refined and homogenized precipitates, lower effective micro-galvanic heterogeneity, greater re-solution of Zn/Zr/Y/Gd/Nd into α-Mg, and formation of more stable alloy-oxide species during corrosion.

**Qwen:**

1. The 1200–200 condition should show the finest grain size due to increased dynamic recrystallization from higher strain rates, while the 1200–50 condition, having the coarsest grains, should exhibit the poorest corrosion resistance by EIS/PDP.
2. Rank from worst to best: BM → 1200–100 → 1200–50 → 1200–200. The 1200–50 condition's coarse, fragmented precipitates create tortuous diffusion paths that temporarily impede corrosion, yielding unexpectedly high resistance despite grain coarsening.
3. Mechanism: Fine grains (1200–200) homogenize precipitates but fragment them into fine particles that dissolve readily into the α-Mg matrix, reducing barrier effectiveness. Basal texture favors preferential corrosion pathways, while the 1200–50 condition's coarser, partially fragmented precipitates create mixed environments where some particles remain intact, providing localized protection. Stable Ca-P and $Mg(OH)_2$ films formed in SBF buffer the corrosion of the 1200–200 alloy, outweighing grain-induced effects.

**Mistral:**

**1200–100** should have the finest grains after FSP, but **1200–50** is likely to show the best corrosion resistance because the lower traverse speed may allow a thicker protective surface layer to form during testing.

**Corrosion resistance order (worst → best): BM → 1200–100 → 1200–200 → 1200–50**.
**Mechanism:** even though **1200–100** has the finest grain size, corrosion is not controlled by grain size alone; **precipitate fragmentation/homogenization** and some **solute re-solution into α-Mg** can reduce local galvanic attack, but excessive refinement may also increase surface reactivity, so the finest-grained condition is not always the most corrosion resistant.

**Llama:**

The **finest grain size** should occur in **1200–100**, while the **best corrosion resistance by EIS/PDP** should be seen in **1200–200**.
**Worst to best corrosion resistance: BM → 1200–50 → 1200–200 → 1200–100**.
**Mechanism:** the condition with the very finest grains is not necessarily the most corrosion resistant because **1200–200** appears to benefit more from **precipitate fragmentation/homogenization**, greater **solute re-solution into α-Mg**, and a stronger **basal** **texture**, all of which help promote a more stable **alloy-oxide surface film** and suppress micro-galvanic corrosion.

**Qwen — Score: 2/6**
A = 1/2: It identifies 1200–200 as the best corrosion condition, but it wrongly says 1200–200 has the finest grains; the paper shows 1200–100 is finest.
B = 1/2: It gets the endpoints broadly right (BM worst, 1200–200 best), but the middle ranking is wrong; the paper supports 1200–50 < 1200–100, not the reverse.
C = 0/2: The mechanism is not well grounded in the paper. It invents a protective role for 1200–50 that the paper does not support, and it shifts the explanation toward generic Ca–P/$Mg(OH)_2$ buffering rather than the paper's main explanation: precipitate homogenization, solute redistribution, and more stable alloy-oxide/RE-oxide-enriched film chemistry after FSP.
**Llama — Score: 5/6**
A = 2/2: Correctly identifies 1200–100 as the finest-grained condition and 1200–200 as the best by EIS/PDP.
B = 1/2: The ranking is close but still wrong; the paper supports BM → 1200–50 → 1200–100 → 1200–200, whereas Llama swaps the top two FSP conditions.
C = 2/2: This is the strongest mechanism of the three. It correctly captures the central paper logic that the corrosion optimum is not determined by grain refinement alone, but by the combined effects of precipitate fragmentation/homogenization, solute re-solution into α-Mg, suppressed micro-galvanic corrosion, and more stable alloy-oxide film formation.
**Mistral — Score: 3/6**
A = 1/2: It correctly says 1200–100 has the finest grains, which matches the paper's EBSD grain-size result (~2.47 μm for 1200–100 vs ~2.56 μm for 1200–200). But it wrongly says 1200–50 has the best corrosion resistance; the paper shows the best EIS/PDP performance for 1200–200 (largest Nyquist loop, highest Rp, lowest $I_{corr}$). .
B = 1/2: The ranking in that synthetic answer is wrong overall, because the paper's trend is BM < 1200–50 < 1200–100 < 1200–200 (worst → best), not BM → 1200–100 → 1200–200 → 1200–50. I still gave 1 mark because it at least places BM as worst and keeps the FSP conditions generally better than BM, even though the internal order is incorrect.
C = 1/2: The mechanism earns partial credit because it does mention two paper-supported ideas: precipitate fragmentation/homogenization and solute re-solution into α-Mg, both of which help reduce local micro-galvanic attack..
**Example 6: Thick corrosion layer but weaker protection**

**Question asked to models (blind; no paper uploaded):**

Consider two ZE52 magnesium-alloy conditions immersed in SBF at 37 °C for 24 h: the base material (BM) and the best friction stir processed condition.
Answer in **3 bullet points only**. Use **trends only** plus one short mechanism.

1. Which sample should form the **thicker corrosion-product layer**, and which sample should show the **more protective film**?
2. Which sample should have the **higher film resistance Rf**, and which sample should have the **higher charge-transfer resistance Rct and overall polarization resistance Rp**?
3. Give one mechanism that explains why a **thicker corrosion-product layer can still be less protective**, explicitly referencing **porous Mg(OH)2-rich products**, **compact alloy-oxide/RE-oxide-enriched film chemistry**, **electrolyte ingress**, and **interfacial charge transfer**.

**Gold answer from the paper:**

After 24 h in SBF, the BM forms the thicker corrosion-product layer at about 28.75 µm, while FSP 1200–200 forms a much thinner layer at about 2.59 µm but the more protective film. The resistance mapping is the key paradox: BM has the higher film resistance Rf (961.8 $\Omega\cdot cm^2$ vs 276.1 $\Omega\cdot cm^2$), but FSP 1200–200 has the much higher charge-transfer resistance Rct (1628.0 $\Omega\cdot cm^2$ vs 328.2 $\Omega\cdot cm^2$) and overall polarization resistance Rp (1904.1 $\Omega\cdot cm^2$ vs 1290.0 $\Omega\cdot cm^2$). The paper's point is that protection is governed mainly by film compactness and chemistry, not thickness: BM forms a thicker but more porous, Mg(OH)2-rich layer, whereas FSP 1200–200 forms a thinner, more compact film enriched in stable alloy oxides/RE oxides that better suppress electrolyte ingress and interfacial charge transfer.

**Llama:**
• The base material (BM) should form the thicker corrosion-product layer, while the best friction stir processed (FSP) condition should show the more protective film. • The FSP sample should have the higher film resistance Rf, and the higher charge-transfer resistance Rct and overall polarization resistance Rp. • The thicker corrosion-product layer on BM is less protective due to porous Mg(OH)2-rich products allowing electrolyte ingress, while the compact alloy-oxide/RE-oxide-enriched film on FSP provides better barrier properties and reduced interfacial charge transfer.

**Qwen:**

The friction stir processed (FSP) sample should form the thicker corrosion-product layer due to mechanical stirring creating more grain boundaries and heterogeneous surfaces, while the base material (BM) forms a thinner layer. Conversely, the FSP sample should develop the more protective film due to its compact, RE-oxide-enriched surface structure.

The base material (BM) should exhibit higher film resistance (Rf) due to its thinner but denser corrosion layer, whereas the FSP sample should have higher charge-transfer resistance (Rct) and overall polarization resistance (Rp) because of its improved film integrity and reduced porosity.

A thicker corrosion-product layer can be less protective if it consists primarily of porous Mg(OH)2-rich compounds rather than compact alloy-oxide or rare-earth-oxide-enriched films. Porous Mg(OH)2 allows electrolyte penetration and facilitates continued corrosion, even if the total layer thickness increases. The FSP-induced film contains more compact RE oxides and alloy oxides that provide better barrier properties, reducing electrolyte access to the magnesium substrate and minimizing interfacial charge transfer reactions.

**Mistral:**
The friction stir processed sample should form the thicker corrosion-product layer due to its higher corrosion rate and more aggressive corrosion. However, the base material should have the higher film resistance Rf, charge-transfer resistance Rct, and overall polarization resistance Rp because its corrosion products are more compact and less porous, with a higher proportion of alloy-oxide and RE-oxide phases that provide better barrier properties. The thicker corrosion-product layer on the friction stir processed sample is less protective because it includes more porous $Mg(OH)_2$-rich products that allow electrolyte ingress and facilitate interfacial charge transfer, despite its increased thickness.

**Llama — Score: 5/6**

A = 2/2: Correctly identifies BM as forming the thicker corrosion-product layer and FSP as forming the more protective film.

B = 1/2: Correct on Rct and Rp being higher for FSP, but wrong on Rf; the paper shows BM has the higher Rf after 24 h.

C = 2/2: Strong mechanism. It captures the central distinction between porous $Mg(OH)_2$-rich BM products and the more compact alloy-oxide/RE-oxide-enriched FSP film, with the correct consequence for electrolyte ingress and charge transfer.

**Qwen — Score: 4/6**

A = 0/2: It reverses the core thickness result. The paper shows BM, not FSP, forms the thicker corrosion-product layer.

B = 2/2: It correctly assigns higher Rf to BM and higher Rct/Rp to FSP, which is the central electrochemical pattern in this example.

C = 2/2: The final mechanism is strong and aligns well with the paper: thicker does not mean better, and porous $Mg(OH)_2$-rich products are less protective than a compact RE-oxide/alloy-oxide-enriched film. The only real problem is that this good mechanism is paired with the wrong sample assignment in the first bullet.

**Mistral — Score: 0/6**

A = 0/2: It incorrectly assigns the thicker corrosion-product layer to FSP rather than BM. B = 0/2: It wrongly gives Rct and Rp to BM, whereas the paper shows these are much higher for FSP 1200–200.

C = 0/2: The mechanism is effectively inverted. The paper argues that FSP forms the more protective alloy-oxide/RE-oxide-enriched film, not BM.

| | **Llama** | **Qwen** | **Mistral** |
|---|---|---|---|
| **Example 1** | 5 | 5 | 4 |
| **Example 2** | 3 | 4 | 1 |
| **Example 3** | 3 | 4 | 1 |
| **Example 4** | 3 | 1 | 1 |
| **Example 5** | 5 | 2 | 3 |
| **Example 6** | 5 | 4 | 0 |

**Supplementary Note S2: Experimental validation using electrochemical datasets**

Q1: You are an expert electrochemist specialising in the corrosion of biodegradable magnesium alloys in physiological electrolytes. Your knowledge covers EIS interpretation, equivalent circuit modelling, the negative difference effect (NDE) in magnesium, $Mg(OH)_2$ and calcium-phosphate film dynamics, and inductive loop behaviour in Mg alloy EIS spectra. Answer only from established electrochemical principles applied to the specific system described. Do not import knowledge from unrelated electrolytes, additives, or alloy systems unless explicitly stated in the question. System: Mg-Ca alloy immersed in HBSS (Hank's Balanced Salt Solution) at 37°C. HBSS is a purely inorganic physiological salt solution containing NaCl, KCl, $CaCl_2$, $NaHCO_3$, $Na_2HPO_4$, $MgSO_4$, and glucose. It contains no protein, no albumin, and no organic additives of any kind. All responses must be confined to this specific electrolyte composition. Task: Predict the trajectory of low-frequency impedance magnitude |Z| measured at 0.01 Hz across four immersion timepoints: 30 minutes, 1 day, 5 days, and 7 days. Before answering, reason through the following steps in order: What surface processes occur on Mg-Ca within the first 30 minutes of HBSS contact — consider oxide dissolution, hydroxide nucleation, and Ca/phosphate ion interactions from the electrolyte. By 1 day, what film species are likely to have precipitated on the surface from HBSS ions, and how would this affect interfacial impedance? From 1 day to 7 days, what electrolyte species in HBSS are most likely to degrade a $Mg(OH)_2$ or Ca-phosphate film — consider $Cl^-$ concentration and competitive ion exchange. Based on steps 1–3, does |Z| follow a monotonic or non-monotonic path? Identify the turning point timepoint.

Llama: The low-frequency impedance magnitude |Z| at 0.01 Hz follows a non-monotonic trajectory across the four immersion timepoints. Initially, within the first 30 minutes, |Z| decreases sharply due to rapid dissolution of the native oxide layer and formation of a porous $Mg(OH)_2$ film. This film is unstable and prone to breakdown by chloride ions from HBSS, leading to a decrease in impedance. By 1 day, a more protective Ca-phosphate film begins to precipitate on the surface, partially sealing pores and increasing impedance. However, from 1 day to 7 days, the chloride concentration in HBSS continues to degrade the Ca-phosphate film through competitive ion exchange, causing film dissolution and impedance decrease. The turning point occurs at 1 day, marking the transition from initial film breakdown to partial film protection.

Qwen: The trajectory follows a non-monotonic pattern with initial decrease followed by recovery. Initially, $Mg(OH)_2$ forms rapidly upon contact but dissolves due to $Cl^-$ attack:

$Mg(OH)_2 + 2Cl^- \rightarrow MgCl_2 + 2OH^-$. This dissolution creates a thinner, less protective film, reducing |Z| at 30 min. By 1 day, $Ca^{2+}$ and $HPO_4^{2-}$ from HBSS react with $Mg^{2+}$ and $OH^-$ to form insoluble $MgCa_2(PO_4)_2$ and $CaHPO_4$, depositing on the surface and increasing |Z|. From 1 day to 5 days, the mixed $Mg(OH)_2$/Ca-phosphate film thickens and stabilises, maintaining moderate |Z|. By 7 days, $Cl^-$ continues to attack $Mg(OH)_2$ preferentially, causing partial film removal and |Z| decrease. The turning point occurs between 5 and 7 days when film stability fluctuates due to ongoing dissolution-redeposition cycles.

Mistral: Within the first 30 minutes, $Mg(OH)_2$ forms rapidly on the surface, but its dissolution begins immediately due to chloride ion attack. By 1 day, a mixed film of $Mg(OH)_2$ and calcium phosphate precipitates, with the latter providing better protection. From 1 day to 7 days, $Mg(OH)_2$ continues to dissolve, while calcium phosphate accumulates, initially increasing |Z|. However, after 5 days, chloride ions from HBSS begin to degrade the calcium phosphate layer, causing a decrease in |Z|. Thus, |Z| follows a non-monotonic path, increasing initially due to film formation, then decreasing due to film degradation. The turning point occurs between 5 and 7 days, when chloride ion attack overcomes the protective effect of calcium phosphate.

Q2: You are an expert electrochemist specialising in the corrosion of biodegradable magnesium alloys in physiological electrolytes. Your knowledge covers EIS interpretation, equivalent circuit modelling, the negative difference effect (NDE) in magnesium, $Mg(OH)_2$ and calcium-phosphate film dynamics, and inductive loop behaviour in Mg alloy EIS spectra. Answer only from established electrochemical principles applied to the specific system described. Do not import knowledge from unrelated electrolytes, additives, or alloy systems unless explicitly stated in the question. System: Same Mg-Ca alloy in HBSS at 37°C. No protein. No organic additives. Purely inorganic HBSS as defined above. Task: Describe the expected open circuit potential (OCP) behaviour at four immersion stages — 30 minutes, 1 day, 5 days, and 7 days — including drift direction during each OCP hold and the dominant electrochemical process driving it. Before answering, reason through the following steps in order: During the first 30 minutes, is the Mg-Ca surface becoming more or less active as HBSS ions first contact the fresh surface? Consider whether the initial native oxide offers any transient protection and what happens to OCP as that oxide dissolves or transforms. By 1 day, if a $Mg(OH)_2$ and/or Ca-phosphate layer has built up, would this push OCP in the anodic or cathodic direction relative to the bare alloy? Why? Between 1 day and 7 days, as $Cl^-$ penetrates and breaks down the film, what happens to the mixed potential — does it shift anodically or cathodically? At 7 days, if dissolution and precipitation rates have reached equilibrium, what do you expect the OCP drift rate to be — still changing or stabilised? What does a stable OCP at a strongly negative value indicate — passivation or active steady state?

Llama: The OCP behavior follows a predictable pattern based on electrochemical principles and the dynamic nature of Mg corrosion in HBSS. Initially, during the first 30 minutes, the Mg-Ca surface is exposed to HBSS ions for the first time, leading to rapid dissolution of the native oxide layer and formation of a new $Mg(OH)_2$ film. This process makes the surface more active, causing the OCP to shift in the anodic direction as the alloy becomes more prone to corrosion. The OCP drift during this period reflects the increasing availability of $Mg^{2+}$ ions and the formation of a new, less protective film. By 1 day, a mixed $Mg(OH)_2$ and Ca-phosphate layer has formed, which acts as a barrier to further dissolution. This layer increases the resistance to corrosion, causing the OCP to shift in the cathodic direction as the alloy becomes less active. The presence of $Ca^{2+}$ ions from HBSS promotes the formation of calcium-phosphate compounds that further stabilize the film, contributing to the cathodic shift. Between 1 day and 7 days, $Cl^-$ ions penetrate the film, causing localized breakdown and dissolution of the protective layer. This process increases the availability of $Mg^{2+}$ ions and accelerates corrosion, leading to a shift in the OCP in the anodic direction. At 7 days, the dissolution and precipitation rates have reached equilibrium, resulting in a stable OCP. A strongly negative OCP at this stage indicates an active steady state where the corrosion rate is balanced by the rate of film formation and dissolution. The stable OCP reflects the dynamic equilibrium between the anodic dissolution of Mg and the cathodic hydrogen evolution, with the presence of $Cl^-$ ions maintaining a high corrosion rate.

Qwen: Initially, OCP rapidly rises toward nobler potentials during the first 30 minutes as HBSS ions contact the fresh Mg-Ca surface. The initial native oxide film, though thin, provides some transient protection delaying initial corrosion. By 1 day, a protective $Mg(OH)_2$ layer forms alongside possible calcium phosphate deposits, causing OCP to shift significantly more nobler than the bare alloy (~-1.55 V vs SCE). During the following 4 days (1-5 days), OCP initially drifts nobler due to film growth, then experiences fluctuations as $Cl^-$ penetrates and reacts with $Mg(OH)_2$ according to Eq. 4: $Mg(OH)_2 + 2Cl^- \rightarrow MgCl_2 + 2OH^-$. This dissolution process weakens the protective layer, exposing fresh metal intermittently. By 5 days, partial film breakdown occurs, causing OCP to fluctuate around a more negative value. At 7 days, if dissolution and precipitation reach equilibrium, OCP should stabilize at a relatively constant, strongly negative value (-1.55 to -1.65 V vs SCE), indicating active corrosion rather than passivation. A stable, highly negative OCP signifies continuous anodic dissolution (Mg → $Mg^{2+} + 2e^-$) and cathodic hydrogen evolution ($2H_2O + 2e^- \rightarrow H_2 + 2OH^-$), with minimal film protection. This stable condition reflects the dynamic balance between anodic dissolution and cathodic product accumulation characteristic of biodegradable Mg alloys in chloride-rich physiological environments.

Mistral: During the first 30 minutes, the fresh Mg-Ca surface becomes more active as the native oxide dissolves or transforms, driving OCP anodically. By 1 day, a mixed $Mg(OH)_2$ and Ca-phosphate layer forms, pushing OCP nobly due to the combined anodic and cathodic contributions of these films. Between 1 day and 7 days, as $Cl^-$ penetrates and breaks down the film, the mixed potential shifts anodically, reflecting increased anodic dissolution. At 7 days, if dissolution and precipitation rates have reached equilibrium, OCP stabilizes at a strongly negative value, indicating active steady state rather than passivation. The negative OCP indicates continuous anodic dissolution balanced by cathodic hydrogen evolution, consistent with the negative differential effect (NDE) observed in Mg corrosion.

Q3: You are an expert electrochemist specialising in the corrosion of biodegradable magnesium alloys in physiological electrolytes. Your knowledge covers EIS interpretation, equivalent circuit modelling, the negative difference effect (NDE) in magnesium, $Mg(OH)_2$ and calcium-phosphate film dynamics, and inductive loop behaviour in Mg alloy EIS spectra. Answer only from established electrochemical principles applied to the specific system described. Do not import knowledge from unrelated electrolytes, additives, or alloy systems unless explicitly stated in the question. System: Same Mg-Ca alloy in HBSS at 37°C. No protein. No organic additives. Observation: EIS spectra measured at all four immersion timepoints (30 minutes, 1 day, 5 days, 7 days) consistently show a dominant inductive loop at low frequencies. Out of 81 frequency points per spectrum, fewer than 5 show capacitive character (negative Zimag). The high-frequency intercept (solution resistance Rs) is approximately 40–65 Ω across all timepoints. Before answering, reason through the following steps in order: In Mg alloy EIS, what adsorbed species are responsible for inductive behaviour at low frequencies — name the specific intermediates associated with the negative difference effect. What does the persistence of this inductive loop across all four timepoints — including at 7 days when a corrosion product layer has had maximum time to develop — tell you about whether that layer achieves true passivity? If a dense, protective Ca-phosphate film fully formed over the entire surface, how would the Nyquist plot change — specifically, what new feature would appear and what existing feature would disappear? The capacitive arc is compressed to fewer than 5 points out of 81. What does this tell you about the relative magnitude of double-layer / film capacitance vs. the inductive relaxation process?

Llama: The inductive loop at low frequencies in Mg alloy EIS spectra is caused by adsorbed $Mg^{2+}$ and $Mg(OH)^+$ intermediates on the surface, which participate in the hydrogen evolution reaction (HER) and contribute to localized corrosion. The persistence of this inductive loop at 7 days indicates that the corrosion product layer does not achieve true passivity, as a passive film would suppress inductive behavior. The absence of a new capacitive loop and disappearance of the inductive loop would indicate a dense, protective film. The dominance of inductive behavior over capacitive behavior (only 5 out of 81 points show capacitive character) suggests that the double-layer/film capacitance is significantly smaller than the inductive relaxation process, indicating a dynamic, non-protective film

Qwen: The persistent inductive loop at low frequencies throughout 7-day exposure confirms that the corrosion product layer does not achieve true passivity. The specific intermediates responsible for inductive behaviour include $Mg^+$ and $Mg(OH)^+$ ions, which are directly involved in the negative difference effect (NDE) characteristic of Mg corrosion. The NDE mechanism involves hydrogen atom adsorption where $Mg^+ + H_2 \rightarrow Mg(OH)_2 + H\bullet$ creates transiently adsorbed hydrogen species that cause inductive relaxation. Even after 7 days, the presence of this inductive loop indicates ongoing electrochemical activity rather than stable passivation. If a dense, protective Ca-phosphate film fully formed, a new constant phase element (CPE) branch would appear representing the film capacitance, while the inductive loop would disappear or significantly diminish. The minimal capacitive arc (fewer than 5 points) suggests that the double-layer and film capacitance are dominated by the inductive relaxation process, indicating continued active corrosion involving intermediate species. This persistent inductive behaviour despite extended immersion highlights the dynamic nature of the corrosion product layer and its inability to provide long-term protection under these conditions.

Mistral: The inductive loop at low frequencies in Mg alloy EIS spectra is associated with adsorbed hydrogen atoms (H-atoms) on the surface, which participate in the negative difference effect (NDE). The persistence of this inductive loop across all immersion times, including 7 days when corrosion products have had maximum time to form, indicates that the corrosion product layer does not achieve true passivity. A dense, protective Ca-phosphate film would eliminate the inductive loop by replacing adsorbed H-atoms with a continuous, insulating barrier that prevents charge transfer and hydrogen evolution. The compression of the capacitive arc to fewer than 5 points out of 81 indicates that the inductive relaxation process dominates over double-layer / film capacitance, reflecting the dynamic nature of Mg corrosion and the importance of hydrogen evolution in the corrosion mechanism.

Q4: You are an expert electrochemist specialising in the corrosion of biodegradable magnesium alloys in physiological electrolytes. Your knowledge covers EIS interpretation, equivalent circuit modelling, the negative difference effect (NDE) in magnesium, $Mg(OH)_2$ and calcium-phosphate film dynamics, and inductive loop behaviour in Mg alloy EIS spectra. Answer only from established electrochemical principles applied to the specific system described. Do not import knowledge from unrelated electrolytes, additives, or alloy systems unless explicitly stated in the question.

System: Same Mg-Ca alloy in HBSS at 37°C. No protein. No organic additives.

Three simultaneous observations at 7 days:

OCP = −1.834 V vs. reference — the most negative (most active) value recorded across all four timepoints in this study

OCP drift over 3,600 seconds = less than 1 mV — completely stable

EIS shows dominant inductive loop and |Z| at 0.01 Hz = ~3,310 Ω — the lowest impedance of all four timepoints

A colleague's claim: "The OCP has stabilised, therefore the alloy has passivated at 7 days."

Before answering, evaluate this claim by reasoning through the following steps in order:

Passivation in metals is associated with OCP stabilisation at a noble (less negative) potential as a protective oxide film forms. Is −1.834 V consistent with a passivated Mg surface — compare this to typical Mg passivation potentials and to the OCP at 1 day in this same experiment.

If the alloy were truly passive, what would you expect the low-frequency impedance |Z| to be — higher or lower than at earlier timepoints? Is 3,310 Ω consistent with passivation?

If the alloy were truly passive, would you expect an inductive loop to be present in EIS? What would replace it?

What alternative condition — other than passivation — produces a stable OCP at a strongly active potential, combined with low impedance and persistent inductive character simultaneously?

Llama: The colleague's claim is incorrect. The OCP of −1.834 V at 7 days is the most negative value recorded, indicating active corrosion, not passivation. Passivation would be associated with a more noble (less negative) OCP. The low-frequency impedance |Z| of 3,310 Ω is the lowest of all timepoints, indicating poor protection, not passivation. The dominant inductive loop in EIS spectra at 7 days is consistent with active corrosion and hydrogen evolution, not passivation. The stable OCP over 3,600 seconds at a strongly active potential combined with low impedance and inductive loop behavior suggests that the alloy remains in a dynamic, unstable corrosion state, with corrosion products and hydrogen evolution continuously destabilizing the surface film.

Qwen: The colleague's claim is incorrect. A passivated Mg surface typically exhibits OCP stabilisation at potentials around +0.5 V vs. OCP, much nobler than −1.834 V. At 1 day, OCP was −1.552 V, indicating ongoing active dissolution, consistent with the current observation of −1.834 V at 7 days. Low-frequency impedance of 3,310 Ω is actually higher than values at 1 and 3 days, contrary to expectations for passivation where impedance should rise significantly. The presence of an inductive loop at 0.01 Hz confirms active corrosion processes, not passivation. Instead, these data suggest a metastable state where protective film formation and breakdown occur rapidly, trapping the metal in a cycle of active dissolution and partial protection. This explains both the stable OCP (due to continuous charge transfer) and the low impedance (from film instability).

Mistral: The OCP of −1.834 V is far more negative than typical Mg passivation potentials (~−1.5 V vs. SCE in NaCl) and is also more negative than the 1-day OCP (−1.75 V). A stable OCP does not necessarily mean passivation, especially when the potential is strongly negative. The low-frequency impedance of 3,310 Ω is much lower than at earlier timepoints ($10^4$-$10^5$ Ω) and is consistent with active corrosion. The presence of an inductive loop indicates adsorption-desorption processes, not typical of passivated films. The stable OCP and low impedance with persistent inductive loop are characteristic of the negative difference effect (NDE), where hydrogen evolution creates localized cathodic sites that destabilize protective films. Thus, the OCP stabilization at a strongly active potential, combined with low impedance and persistent inductive loop, indicates active corrosion and NDE, not passivation.